\pdfoutput=1
\documentclass[10pt,twocolumn,letterpaper]{article}

\usepackage[pagenumbers]{iccv} % To force page numbers, e.g. for an arXiv version

\usepackage{csquotes}
\usepackage{soul}
\usepackage{multirow}

\usepackage[utf8]{inputenc}  % For older LaTeX versions
\usepackage[T1]{fontenc}      % Improved font encoding

\DeclareUnicodeCharacter{2714}{\checkmark}
\DeclareUnicodeCharacter{FFFD}{}

\definecolor{pltorange}{HTML}{ff7f0e} 
\definecolor{pltblue}{HTML}{1f77b4} 
\definecolor{greenjesus}{rgb}{0.01, 0.75, 0.24}

\newcounter{magicrownumbers}
\newcommand\rownumber{\stepcounter{magicrownumbers}\scriptsize{\arabic{magicrownumbers}}}

\usepackage{pifont}
\DeclareUnicodeCharacter{2718}{\ding{55}}

\definecolor{iccvblue}{rgb}{0.21,0.49,0.74}
\usepackage[pagebackref,breaklinks,colorlinks,allcolors=iccvblue]{hyperref}

\usepackage{comment}

\def\paperID{12798} % *** Enter the Paper ID here
\def\confName{ICCV}
\def\confYear{2025}

\title{Conjuring Positive Pairs for Efficient Unification of \\Representation Learning and Image Synthesis}
\author{
Imanol G. Estepa\footnotemark[1]\\
Universitat de Barcelona, Spain\\
\href{mailto:estepa.gonzalez@ub.edu}{\tt\small estepa.gonzalez@ub.edu}
\and
Jesús M. Rodríguez-de-Vera\thanks{Equal contribution}\\
Universitat de Barcelona, Spain\\
\href{mailto:j.molina.rdv@ub.edu}{\tt\small j.molina.rdv@ub.edu} % Make email clickable
\and
Ignacio Sarasúa\\
NVIDIA Computing Spain\\
\href{mailto:isarasua@nvidia.com}{\tt\small isarasua@nvidia.com}
\and
Bhalaji Nagarajan\\
Barcelona Supercomputing Center (BSC)\\
\href{mailto:bhalaji.nagarajan@bsc.es}{\tt\small bhalaji.nagarajan@bsc.es}
\and
Petia Radeva\\
Universitat de Barcelona, Spain\\
\href{mailto:petia.ivanova@ub.edu}{\tt\small petia.ivanova@ub.edu}
}

\begin{document}
\maketitle
\begin{abstract}

While representation learning and generative modeling seek to understand visual data, unifying both domains remains unexplored. Recent Unified Self-Supervised Learning (SSL) methods have started to bridge the gap between both paradigms. However, they rely solely on semantic token reconstruction, which requires an external tokenizer
% . to convert input images into semantic tokens 
during training --- introducing a significant overhead. 
In this work, we introduce \emph{Sorcen}\footnote{The source is available in \href{https://github.com/ImaGonEs/Sorcen}{https://github.com/ImaGonEs/Sorcen}}, a novel unified SSL framework, incorporating a synergic Contrastive-Reconstruction objective. 
Our Contrastive objective, \enquote{Echo Contrast}, leverages the generative capabilities of Sorcen, eliminating the need for additional image crops or augmentations during training. 
Sorcen \enquote{generates} an echo sample in the semantic token space, forming the contrastive positive pair.
% \ima{Maybe this we can remove: From each input, Sorcen generates an Echo sample, which is then used to form the contrastive pair. Unlike previous unified state-of-the-art methods, this contrastive loss not only preserves but enhances the model's image synthesis performance, in addition to improving representation learning.} 
% Building on this, 
Sorcen operates exclusively on precomputed tokens, eliminating the need for an online token transformation during training, thereby significantly reducing computational overhead. 
Extensive experiments on ImageNet-1k demonstrate that Sorcen outperforms the previous Unified SSL SoTA by \textbf{0.4\%}, \textbf{1.48 FID}, \textbf{1.76\%}, and \textbf{1.53\%} on linear probing, unconditional image generation, few-shot learning, and transfer learning, respectively, while being \textbf{60.8\%} more efficient. 
Additionally, Sorcen surpasses previous single-crop MIM SoTA in linear probing and achieves SoTA performance in unconditional image generation, highlighting significant improvements and breakthroughs in Unified SSL models.

\end{abstract} 
\section{Introduction}

Self-Supervised Learning (SSL) and Image Synthesis are two prominent areas in Computer Vision \cite{hondru_masked_2025, ozbulak_know_2023, chen_progress_2024, yang_diffusion_2023}. 

\begin{figure}[t]
    \centering
    \includegraphics[width=\linewidth]{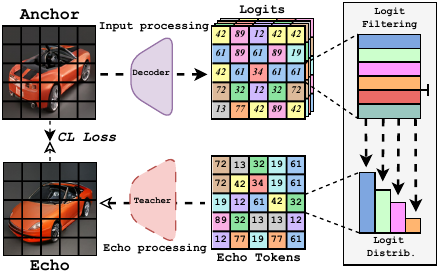}
    \caption{
    \textbf{Simplified Echo Contrast.} For a given anchor input, Sorcen's decoder outputs a set of logits, that are filtered to create a logit distribution, from which diverse Echo tokens are sampled. These tokens are processed by a Teacher encoder before contrasting them against the anchor. Echo contrast enables single input contrastive learning, removing the need of pixel augmentations.}
%    \todo{Update Extension->Expansion}
    \label{fig:echos}
\end{figure}

While both % research areas 
aim to understand and represent visual data, they operate in fundamentally different ways and remain disconnected from each other \cite{hu_complexity_2023}. 
SSL methods use % several 
pretext tasks such as Contrastive Representation learning \cite{zbontar_barlow_2021, chen_exploring_2021, chen_simple_2020, caron_deep_2018} and Masked Image Modeling \cite{he_masked_2022, zhang_how_2022, oquab_dinov2_2023, zhou_ibot_2022} to train general models capable of performing on multiple domains and discriminative downstream tasks. % \rs{thanks to their unsupervised pretext tasks}
%\cite{chen_simple_2020, chen_exploring_2021, caron_deep_2018, zbontar_barlow_2021, oquab_dinov2_2023, zhou_ibot_2022}. \bhalaji{move citation closer to representation learning and mim}
Image synthesis methods \cite{yu_vector-quantized_2021, rombach_high-resolution_2022}, on the other hand, excel in %\rs{leveraging available data to} 
learning distributions that represent the underlying visual information, capable of sampling high-fidelity images on inference when combined with an external model guidance \cite{zhang_towards_2024, yang_diffusion_2023}. 
% \rs{Although each approach}
Both approaches show strong generalization within their respective domains, however, remain bounded to their specific task lines, whether discrimination or image generation.
%Despite each approach demonstrating strong generalization within their own lines, they remain bounded to them, either discrimination tasks or image generation. 

%, \bhalaji{\rs{that can} making it} depict almost anything combined with an external model guidance \cite{zhang_towards_2024, yang_diffusion_2023}.

\begin{figure*}[!ht]
    \centering
    \includegraphics[width=\textwidth]{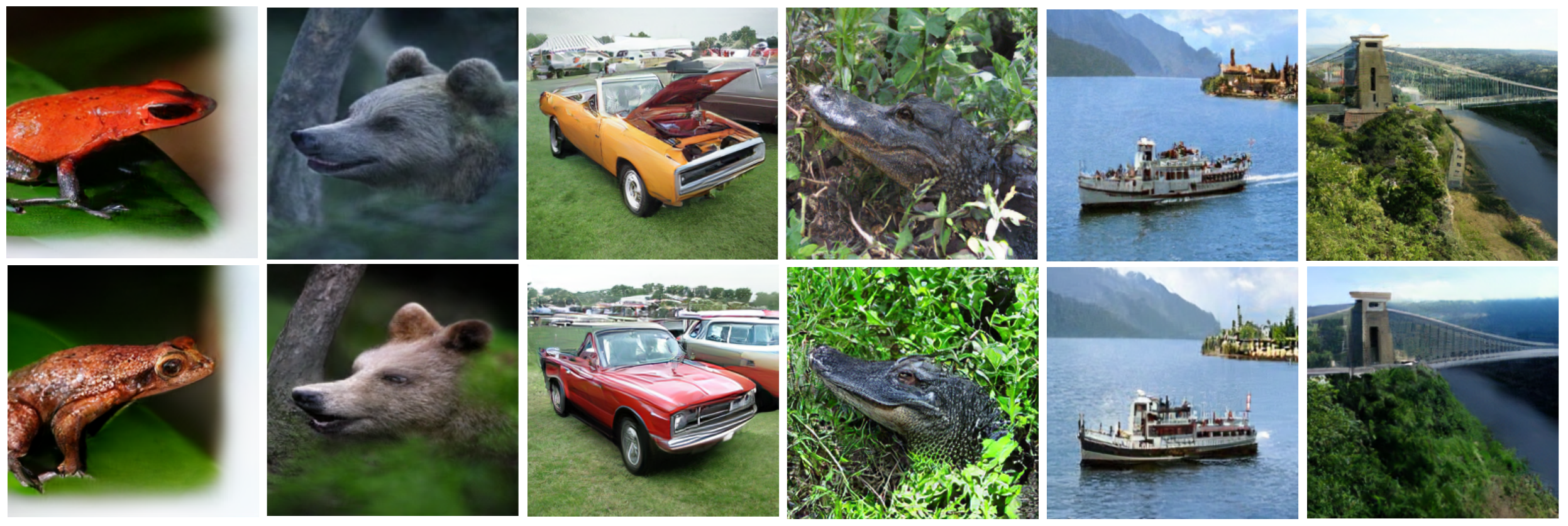}
    \caption{
    \textbf{Echo sample visualization:} 
    Images on top correspond to decoded original token inputs that serve as an \textbf{Anchor} for the \textbf{Echo} samples (bottom). Echoes are extracted during training and decoded on a single decoding step for visualization purposes.}
%    \todo{Update Extension->Expansion}
    \label{fig:echos_fig}
\end{figure*}

% \bhalaji{\rs{Recently} 
Under representation learning, Mask Image Modeling (MIM) % \rs{provided} 
showed very strong % representation learning \bhalaji{we need something different here.} 
discriminative capacity, % while displaying 
with limited image reconstruction capabilities,
%methods managed to display some image reconstruction capabilities while providing a very strong representation learning capacity, 
slightly %bridging 
narrowing the gap between SSL and Image synthesis \cite{he_masked_2022, zhang_how_2022, zhu_stabilize_2024}. 
Contrary to previous multi-crop methods \cite{caron_emerging_2021, zhou_ibot_2022, chen_empirical_2021} in Contrastive % Representation 
learning, MIM methods leverage a single image reconstruction pretext task enabling more efficient frameworks that do not depend on strong augmentations. % augmented pairs. 
Similarly, recent Image synthesis methods \cite{tian_addp_2024, peebles_scalable_2023} begin exhibiting discriminative abilities apart from their usual high-fidelity image generation. 
% Even though these methods 
Although both families of methods support a secondary task alongside their primary function (whether discrimination or generation), their performance remains far from competitive in the other paradigm.
In both lineage, methods often require specialized setups and do not exhibit any symbiotic behavior between the primary and secondary tasks
\cite{bender_dangers_2021}, i.e. the secondary task performance appears to be a consequence rather than an improvement % motivation of 
motivating the overall performance of the model. 
Ideally, a \enquote{unified} model should be able to provide strong performance in both image synthesis and discrimination with an architecture that can leverage the twin strategies, improving both lines \cite{li_mage_2023}. 
Unified SSL is driven by this need, providing models with the capacity to obtain SoTA performance in several discriminative tasks and in unconditional image generation \cite{li_mage_2023}. 
% Still, the reach of these models remain limited as they exclusively rely on semantic token reconstruction pretext task. 
These models exclusively rely on semantic token reconstruction pretext task, limiting their success.
Recent explorations on this pretext task \cite{li_mage_2023} in various setups showed that % further 
improving one of the task lines, either discriminative or generative, often decreased the other. 
% Introducing a Contrastive Representation learning objective, for example, managed to increase the discrimination capacity of the models but dramatically decreased their generation ability \bhalaji{redundant}. 
Furthermore, creating the required semantic token space on-the-fly 
%working on this semantic token space 
adds a big overhead to the already costly pretraining.

On this front, % In this paper, 
% we tackle % encompass 
% several problems detected in the unified SSL SoTA % and elegantly addresses them 
we propose a novel unified architecture, \textbf{Sorcen}, 
% Sorcen presents an innovative architecture that 
leveraging both reconstruction and contrastive pretext tasks. %, addressing the challenges established. 
Our reconstruction, % objective , even if it also 
also based on semantic tokens, is adapted to work with % previously
precomputed tokens instead of computing them online during training, thereby completely removing a substantial overhead. %\bhalaji{okay to say large?}
Different to previous SoTA, MAGE % (MAsked Generative Encoder) 
\cite{li_mage_2023}, our innovative contrastive objective does not require additional % from any 
multi-image setup to improve the discriminative capacity of the model, as it emerges directly from the generative ability of Sorcen. 
% As explained in Figure \ref{fig:echos}, 
Instead of hand-made positive pairs, Sorcen generates % produces 
an \enquote{Echo} sample directly from \enquote{anchors} using the reconstruction objective % from the initial input 
before contrasting both. 
We call this \enquote{Echo contrast} (see Figure \ref{fig:echos}).
The generated Echoes % depicted in Figure \ref{fig:echos_fig}, 
provide the  diversity required to perform a meaningful contrast without any supervision or common augmentations (see Figure \ref{fig:echos_fig}).

Sorcen combines reconstruction and contrastive objectives, creating a synergy that enhances both representation learning and image generation capabilities simultaneously.
We analyze the performance of Sorcen in class unconditional generation and several discriminative tasks. 
% When compared to MAGE, the previous state-of-the-art, 
Compared to MAGE, Sorcen demonstrates superior performance across all tested domains and tasks, while offering improvements in efficiency in terms of memory usage and training time. 
% It also
Sorcen outperforms all single-crop MIM baselines and achieves top results in the unconditional image generation task. 
Our extensive experiments and SoTA performance introduce a novel perspective % position Sorcen as a significant advancement in SSL: 
in unified SSL, % method that improves upon previous unified SoTA, 
while also competing with (and in some cases even surpassing) % earlier 
traditional SSL methods. We summarize our contributions as:
% \begin{itemize}[wide = 0pt]
    % \item 
    
    (1) We propose \textbf{Sorcen}, an innovative unified SSL method that synergistically combines contrastive and semantic reconstruction objectives, surpassing previous SoTA on both representation learning and image synthesis. 
    % \item 
    
    (2) We define a novel Contrastive strategy, called \textbf{Echo Contrast}, that leverages single crop reconstruction capabilities of the model to produce rich and diverse positive samples without any additional crop or pixel augmentations. 
    % \item 
    
    (3) Built over precomputed semantic tokens, Sorcen is % establishes as 
    the \textbf{most efficient unified SSL method}, with $\approx$60\% reductions in training GPU hours compared to existing solutions. % \bhalaji{we need to expand this}
    % \item 
    
    (4) Our \textbf{extensive analysis} highlights the ability of Sorcen, outperforming previous unified SoTA on all tested tasks, % obtaining SoTA performance 
    particularly on unconditional image generation and surpassing previous single crop MIM methods on linear probing.
% \end{itemize}

% Contrastive learning

% Generative models

% Natural step, MIM

% Hybrid SSL
\section{Related Works}

%\bhalaji{If we follow the strategy of related works in PAS, we can avoid in each paragraph telling about our contributions.. (might reduce space this way)...}

\noindent
\textbf{Representation learning in Computer Vision.} 
%Traditional contrastive learning methods \cite{chen_simple_2020, chen_improved_2020, chen_empirical_2021, chen_exploring_2021} have ruled unsupervised representation learning over the last half-decade. Their capacity to pretrain models without relying on any annotation enabled the use of raw image data, providing models able to generalize better than those trained in a supervised fashion \cite{zhou_image_2021}.
Traditional contrastive learning methods \cite{chen_simple_2020, chen_improved_2020, chen_empirical_2021, chen_exploring_2021} have dominated unsupervised representation learning, allowing pretraining without annotations to produce models that generalize better than those trained with supervision \cite{zhou_image_2021}.
% Methods such as 
MoCo methods \cite{chen_improved_2020, chen_empirical_2021} train their models by generating image pairs from a single image using an augmentation pipeline and pulling together different representations.
% obtained from those two pairs. 
At the same time, the nature of popular InfoNCE loss \cite{oord_representation_2019} pushes away the rest of the representations, clustering the latent space without any supervised signal. 
While effective, these approaches suffer from the lack of sample diversity.
% , which has been solved by Innovative strategies 
Methods like neighbor contrast \cite{dwibedi_little_2021, estepa_all4one_2023} and synthetic positive generation \cite{zeng_contrastive_2025} provide innovative strategies to increase this diversity. 
%Still, these strategies remain unreliable or require an external generative model, which increases the overhead of the model. 
Given an anchor image, providing positive pairs with large intra-class variations while still belonging to the same semantic class is a hard task to do, in an uncontrolled and efficient way. 
Sorcen, for the first time, proposes a novel Echo-contrastive strategy that efficiently provides desired positive pairs without relying on external generative models or hand-made pixel augmentations.
% \noindent
SimMIM \cite{xie_simmim_2022}, MAE \cite{he_masked_2022} and BeiT \cite{bao_beit_2021} 
% raised as an alternative to previous 
are an alternative to SSL contrastive learning approaches. 
Instead of relying on image pairs, MIM methods use reconstruction pretext tasks that can train models as good as the contrastive pretext task, but in a more efficient way.
MAE \cite{he_masked_2022}, for example, uses masks up to 75\% of the input images, while working with only the remaining 25\%. 

The increased popularity of transformers \cite{vaswani_attention_2017} and its efficiency led to rapid development of these methods. 
U-MAE \cite{zhang_how_2022} studied the dimensional collapse in MAE and proposed a novel loss to overcome it. 
% Works such as 
I-JEPA \cite{assran_self-supervised_2023} and XTRA \cite{amrani_sample-_2024} present architectures that further improve the efficiency of pretraining. 
CAE \cite{chen_context_2024}, CMAE \cite{huang_contrastive_2024} and iBot \cite{zhou_image_2021} have successfully mixed MIM with contrastive learning strategies.
iBot \cite{zhou_image_2021} leverages % the use of 
multiple image crops and a self-distillation architecture to acquire visual semantics and provide SoTA results. 
CAE \cite{chen_context_2024} and CMAE \cite{huang_contrastive_2024} require only a single sample. 
However, different masking strategies are applied to create positive pairs and, consequently, the same image is contrasted. 
In contrast, Sorcen works by providing a sample that belongs to the same semantic class, however, contains intuitive variations that make it different.

\noindent
\textbf{Unified SSL.}
While some MIM methods \cite{he_masked_2022} % have the capacity to 
reconstruct images, the quality of generated images differs greatly from other generation-only methods. 
% Over the years, 
Generative Adversarial Networks (GANs) are % have proven to be 
strong image generator with high fidelity results \cite{donahue_large_2019, casanova_instance-conditioned_2021, lucic_high-fidelity_2019, liu_diverse_2020}. 
Diffusion models, while more specialized in conditional generation, leverage denoising strategies to provide results on par with GANs \cite{dhariwal_diffusion_2021, rombach_high-resolution_2022}. 
% The popularity of the Transformer architecture has led to improved approaches such as 
HiT \cite{zhao_improved_2021} in GANs and DiT \cite{peebles_scalable_2023} in diffusion  have further enhanced the generation quality. 
MaskGIT \cite{chang_maskgit_2022} %, leveraging the Transformer architecture, 
used a novel generative approach capable of creating high-fidelity and high-resolution images. % Recently, 
% Works such as 
BigBiGAN \cite{donahue_large_2019}, l-DAE \cite{chen_deconstructing_2024} and ADDP \cite{tian_addp_2024} prove that generative features lead to decent image recognition results, opening an interesting line where a single SSL model could excel in both generative and image recognition tasks. 
% Still, 
MAGE \cite{li_mage_2023} remains the most competitive unified SSL model. 
Similarly to token-based discriminative methods \cite{park_seit_2023, lee_seit_2025}, MAGE \cite{li_mage_2023} leverages VQGAN \cite{esser_taming_2021} and proposes a semantic reconstruction objective that manages to train a model for both unconditional generation and discriminative tasks. 
However, VQGAN \cite{esser_taming_2021} introduces a dramatic overhead to the method.
Its contrastive version (MAGE-C), on the other hand, sacrifices generative capabilities to improve discriminative performance. 
Our work, Sorcen, proposes a novel contrastive strategy that allows for the improvement of both generative and discriminative capabilities of the model. 
Its unique Echo-contrast strategy does not require dual-augmented images \cite{zhou_image_2021, chen_empirical_2021}, previously trained generative models \cite{zeng_contrastive_2025} or any additional module \cite{li_mage_2023} such as VQGAN.
Sorcen % provides SoTA results, thereby opening 
contributes to a promising line of efficient single-image contrastive learning.
%and a VQGAN-free pretraining that completely removes the overhead. 

\begin{comment}
  \textbf{Disk efficient learning. } Modern disk efficient methods explore the idea of pretraining SSL models with as many samples as possible while reducing the required disk storage drastically. Sampling strategies such Uniform Random Sampling and C-Score \cite{jiang_characterizing_2021} manage to reduce the required space by reducing the amount of training samples. SeiT \cite{park_seit_2023} and SeiT++ \cite{lee_seit_2025} propose the use of the ViT-VQGAN \cite{yu_vector-quantized_2021} to tokenize datasets and reduce the required disk space. This approach leads to 99\% of reduction in the case of ImageNet-1k dataset and achieve competitive discriminative results. Our method leverages the unsupervised VQGAN \cite{esser_taming_2021} to surpass previous disk efficient methods w.r.t the required size and stands as the first disk efficient hybrid method.  
\end{comment}

\begin{figure*}[!ht]
    \centering
    \includegraphics[width=\textwidth]{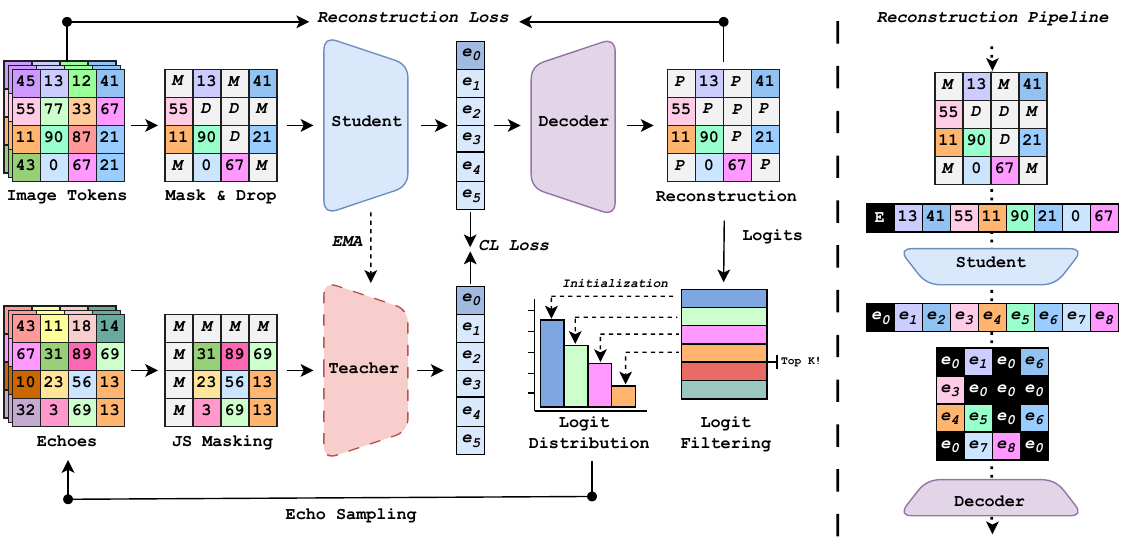}
    \caption{
    \textbf{Left.} Sorcen is conformed by two different objectives: (1) Semantic Reconstruction and (2) Echo Contrast. Semantic reconstruction objective masks (M) and drops (D) the input tokens before processing it by a Student encoder. The Decoder is trained to predict the original input. Using this prediction, Sorcen extracts the logits and use them to sample Echoes, which are the positive samples in Echo Contrast objective. These Echos are processed by a Teacher encoder and the loss is computed between the output % Echos and the previous output 
    of the Student and Teacher encoders. 
    \textbf{Right.} Sorcen leverages an extra token to produce a learnable placeholder embedding for all masked and dropped tokens during the Mask\&Drop phase. This ensures equal token counts for Decoder input/output, which is crucial for Semantic Reconstruction.}
%    \todo{Update Extension->Expansion}
    \label{fig:sorcen_fig}
\end{figure*}

\section{Our Method}

\begin{comment}
 \bhalaji{We should decide if we want to write everything as "we do this, our objective is this..." or if we want to write "sorcen does this, the objective of sorcen is this...." and stick to that uniformly!!!!}   
\end{comment}

Sorcen (see Figure \ref{fig:sorcen_fig}) is a novel unified SSL method capable of both representation learning and unconditional image generation. Sorcen carefully combines a semantic reconstruction objective, which aims to predict the input from a masked version, and a novel contrastive learning objective, which includes \emph{generating} a positive sample, called Echo, before contrasting it against the input. 
%While most contrastive learning approaches utilize hand-made pixel augmentations to create the positive sample in the contrastive pair \cite{chen_empirical_2021, chen_simple_2020, huang_contrastive_2024}, 
%we use the reconstruction capabilities to create echoes, leveraging token space
We use reconstruction capabilities of Sorcen to create Echos, making them diverse and ideal for a contrastive objective. 
% From a single input, our method 
We sample the Echos directly in the semantic token space, making it more efficient than dual-crop pixel-based methods, which additionally require an augmentation pipeline. 
We %This positive sample, defined as \textbf{Echo}, is generated by Sorcen on-the-fly in a much more efficient way than "hand-made" augmentations. As it leverages its own reconstruction capabilities, we 
define this sampling and contrast strategy as Echo contrast, an innovative approach that leverages the advantages of semantic token space to produce and contrast diverse variations of the input or anchor sample.
Sorcen, similar to MAGE \cite{li_mage_2023}, applies its learning objectives in a VQGAN-based \cite{esser_taming_2021} Semantic Token space rather than in a pixel space. 
However, Sorcen computes the semantic tokens only once, removing the overhead of using VQGAN in an online manner.
%, making it more efficient. 
%Sorcen computes distinct semantic tokens for each input image using a VQGAN \cite{esser_taming_2021}.
%which succesfully transform input images into a rich token space. 
%In contrast to MAGE \cite{li_mage_2023}, 
%revious works \cite{li_mage_2023, chang_muse_2023} rely on pixel augmentations and, therefore, require the use of this tokenizer in an online manner, introducing huge overhead to the pretraining phase. 
%Sorcen stands as an improvement, removing this need and  all the overhead involved. 
%The proposed method enables a pixel augmentation-free training that produces meaningful representations for both discriminative and generative tasks. 

%\bhalaji{Can we say in this order (rephrase the sentences to make it better...)
% Two popular architectures can be found in Sorcen. First, it 
Sorcen follows the common Encoder - Decoder architecture for the reconstruction objective, defined as $\mathcal{E}$ and $\mathcal{G}$ respectively. Echo contrast, on the other hand, is performed using a Student-Teacher architecture, where the student is the encoder from the reconstruction objective. The teacher, which is a smoothed version of the student, processes the Echos before the contrastive loss, % and is a smoothed version of the student, 
and is updated every step via Exponential Moving Average (EMA) \cite{zhou_image_2021, grill_bootstrap_2020}. 
This carefully designed architecture enables a pixel-augmentation-free training that produces meaningful representations for both discriminative and generative tasks.

\noindent
\textbf{Input preparation.}
% Sorcen leverages semantic tokens to efficiently produce rich representations. 
Semantic tokens capture the main concepts of images and store them in a smaller space that requires considerably less computational resources to work with. 
%The VQGAN \cite{esser_taming_2021}, while a powerful and effective token generator, comes with a considerable overhead when applied during training. 
%To , 
For every image $\mathcal{I}$ in the dataset $\mathcal{D}$, we precompute the sequence of semantic tokens $\mathbf{t}=(t_1,\dots,t_{CS})$; $CS$ is VQGAN codebook size (tokenizer).  %Each sequence $\mathbf{t}$ contains 256 semantic tokens represented by integers in the range from 0 to 1023. 
%Following MAGE \cite{li_mage_2023}, we use a pretrained VQGAN \cite{esser_taming_2021} with a vocabulary size $v_{max}$ of 1024. 
This vector $\mathbf{t}$ is made up of integers in the range $[0,v_{max}]$, where $v_{max}$ is the tokenizer vocabulary size. % of the tokenizer. 
This set of encoded vectors forms the Sorcen pretraining token dataset. 
Note that token generation phase is carried out once % a single time 
for each dataset and is not attached to Sorcen's pretraining phase, removing 
% This way, we remove 
any possible overhead related to image-token transformation.

\subsection{Semantic Reconstruction Objective}
Our semantic reconstruction objective is divided into: Mask \& Drop phase and Reconstruction objective.

\noindent
\textbf{Step 1: Mask \& Drop.} 
Following typical reconstruction-based methods \cite{li_mage_2023, he_masked_2022, zhou_image_2021}, Sorcen randomly masks its input sequence $\mathbf{t}$, 
% . This sequence will act as 
making it the groundtruth of the objective and Sorcen will be trained to reconstruct it. 
Sorcen follows a truncated Gaussian distribution \cite{he_masked_2022}, where some tokens of every input token sequence are replaced by a constant mask token, $M$, given a binary mask $BM=(m_i)_{i=1}^{CS}$. 
%Following MAGE \cite{li_mage_2023} and MAE \cite{he_masked_2022}, we center the Gaussian distribution at 0.55 and truncate it so only values from 0.5 and 1.0 can be sampled.
The Gaussian distribution is centered at 0.55 and truncated between 0.5 and 1, following MAGE \cite{li_mage_2023} and MAE \cite{he_masked_2022}.
% Once the masking is finished, 
After masking, we further remove or drop half of the masked tokens, 
% decreasing the computational overhead of the model by 
reducing the input size to $L$ tokens \cite{li_mage_2023, he_masked_2022}, thereby decreasing the computational overhead. % of the model.
Finally, an extra token, $E$, is prepended to the sequence, forming a final sequence, $\mathbf{t}'$. 
The extra token is prompted to learn global information about the sequence $\mathbf{t}'$, which will help in reconstruction. % objective. 
At the end of Mask \& Drop, the token sequence, $\mathbf{t}'$ of length $L+1$ is passed through the student encoder, $\mathcal{E}$ projecting the token sequence to the unified latent space. % of Sorcen.

\noindent
\textbf{Step 2: Target prediction.} 
According to the previously applied Mask \& Drop binary masks, the unified latent space
% which is used by our two objectives, 
is adapted for the reconstruction objective. 
Let $\mathcal{E}(\mathbf{t}')=(e_0,e_1,\dots,e_L)$ be the output of the encoder. 
We construct a decoder input sequence $\mathbf{s}=(s_0, s_1,\dots,s_{CS})$ by placing $e_0$ (the output corresponding to the prepended extra token $E$) in the position of every masked or dropped token. 
For the rest of the tokens, defined as visible tokens in reconstruction literature \cite{li_mage_2023, he_masked_2022, huang_contrastive_2024}, the corresponding output $e_i$ is used. 
This transformation process, depicted in Figure \ref{fig:sorcen_fig} (right), perfectly aligns the unified latent with the original token sequence $\mathbf{t}$ and allows proper reconstruction.
%(with position shifts if needed to ensure alignment with the original $\mathbf{t}$).
%In our framework, we use the output of $t_0$, $e_0$, as a placeholder for all dropped and masked tokens, forming a sequence $\mathbf{s}$ made by the outputs of the visible tokens and one $e_0$ for every masked and dropped token. This final sequence $\mathbf{s}$ is reordered according to the original input tokens. 
%This transformation process is depicted in Figure \ref{fig:sorcen_fig} (right). 
Finally, the decoder $\mathcal{G}$ processes $\mathbf{s}$ to predict the original semantic token sequence $\mathbf{t}$.
%starting from $\mathbf{s}$, is encouraged to reconstruct or predict the original semantic token sequence $\mathbf{t}$. 
The reconstruction loss $\mathcal{L}_{\text{Recon}}$ is applied between the original semantic token sequence $\mathbf{t}$ and the output of $\mathcal{G}$, defined as $p$:
\begin{equation*}
 \mathcal{L}_{\text{Recon}} = -\mathbb{E}_{\mathbf{t} \sim \mathcal{D}} \left( \sum_{i=1}^{CS} m_i \log p(t_i | s_i) \right).   
\end{equation*}
Following previous works \cite{he_masked_2022, huang_contrastive_2024, li_mage_2023}, $\mathcal{L}_{\text{Recon}}$ is applied only % exclusively 
to the masked tokens expressed by the binary mask $BM$.

\subsection{Echo Contrast Objective}
The Echo Contrast objective exploits the output of the semantic reconstruction objective to produce Echoes, the target positive samples. It is divided into 3  steps.

\noindent
\textbf{Step 1: Conjuring positive pairs.}
%Different to popular contrastive learning methods \cite{chen_simple_2020, chen_empirical_2021, zhou_image_2021}, 
Sorcen uses the reconstruction capacity to generate useful positive pairs that are beneficial for contrastive learning. 
For every $p$, which contains the prediction logits generated by the decoder $\mathcal{G}$, we initialize a Categorical distribution and sample Echoes from it. 
We control the diversity of Echoes using only the Top-K most probable logits in the distribution. 
%For a given output of the encoder $\mathcal{E}(\mathbf{t})$, our decoder predicts the original input $\mathbf{t}$, generating in the process a vector $\mathcal{p}$ of $v_{max}$ logits (one per possible value) for every token in $\mathbf{s}$. By transforming the logits into a Categorical probability distribution, Sorcen samples the Echos
%diverse variations of $\mathbf{t}$, which we define as Echos, 
%and uses them as positive samples for its contrastive objective. We exclusively use Top-K most probable logits per token to create our Categorical distribution, controlling the diversity of the Echo samples. 
As random sampling could possibly generate an exact copy of the input as an Echo, we always skip the Top-1 logit in our Top-K filtering. 
Figure \ref{fig:echos_fig} shows some generated Echoes. 
%As we work on semantic token space, the overhead added by this process is negligible and enables augmentation-free contrast with just one anchor image. \ima{Improve final phrase}
The sampled Echoes retain relevant information from the anchor and provide intuitive \enquote{distortions} not achievable by common augmentation strategies. 
Since this process is performed in the semantic token space, the overhead is negligible. 

\begin{figure}[t]
    \centering
    \includegraphics[width=\linewidth]{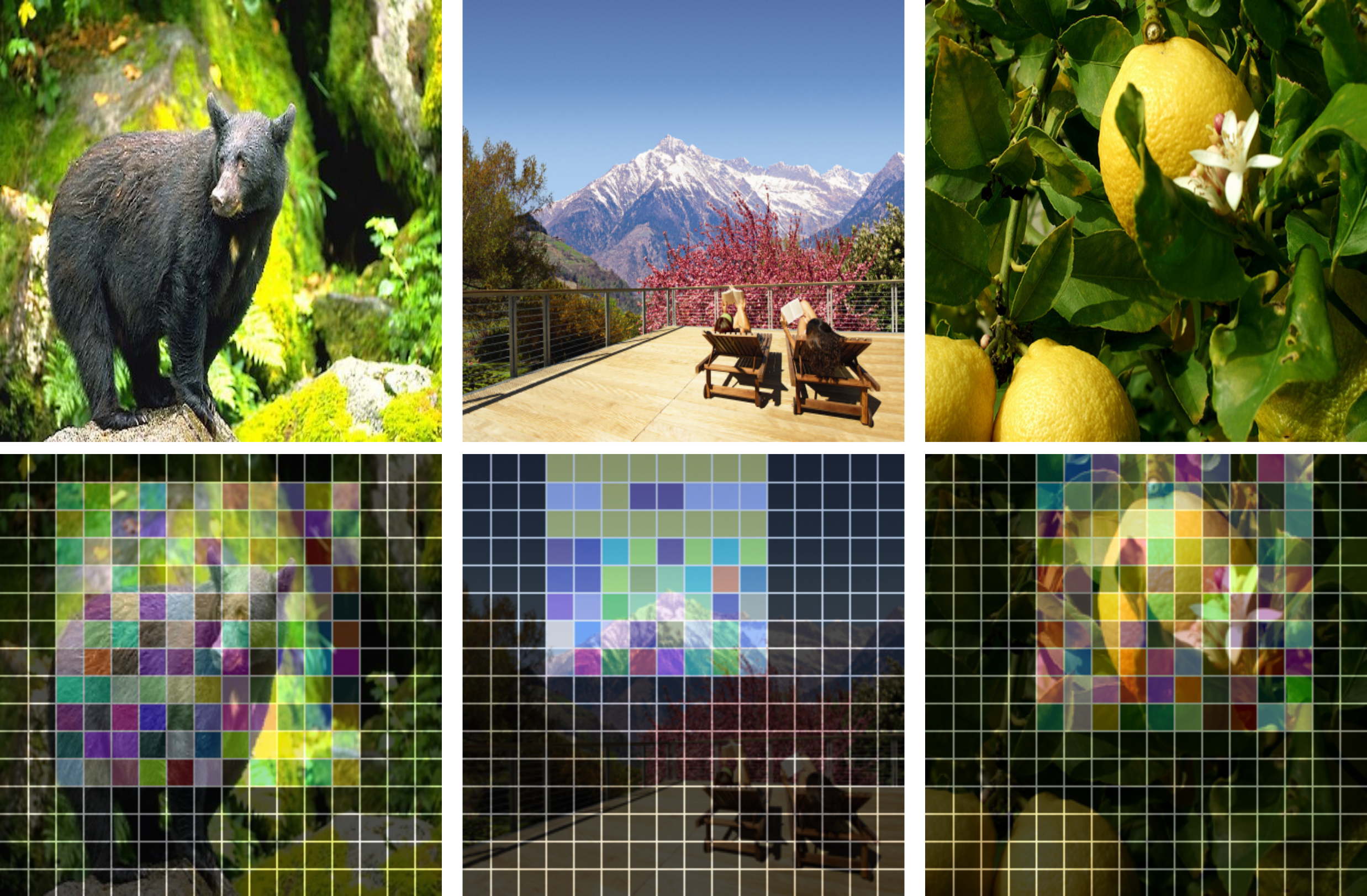}
    \caption{
    JSM visualization. Each cell represents a VQGAN semantic token. Gray cells are the \enquote{cropped-out} area.}
    \label{fig:jsm-main}
\end{figure}

\noindent
\textbf{Step 2: Jittered Spatial Masking (JSM).} 
Inspired by pixel-based works \cite{hinojosa_colormae_2025, huang_contrastive_2024}, we introduce spatial masking to our Echoes, increasing the diversity while providing useful information to the contrast. 
For each Echo, we randomly select a region of size $N \times N$ and mask the rest of the tokens using the $M$ token. 
The location and size of selected region is chosen randomly during training, providing different local information each time. 
This spatial masking, as it maintains a region of the sequence untouched, differs from the one applied in reconstruction objective and provides different information to our contrast objective. 
 Figure \ref{fig:jsm-main} visualizes this masking strategy. Once masked, the Echoes are projected to the unified latent space by our Teacher encoder.

\noindent
\textbf{Step 3: Pulling representations.}
%We leverage a Teacher-Student architecture where the Teacher $\mathcal{E}'$ is a smoothed version of the Student encoder $\mathcal{E}$. 
After the JSM, the Teacher model transforms the Echo samples to obtain the corresponding hybrid latent. 
At this point, the processed sequence $\mathbf{t}'$ and its corresponding Echo have been projected into the unified latent space, enabling the Sorcen's contrastive objective. After applying global average pooling on both representations, InfoNCE \cite{oord_representation_2019} loss is applied. 
The process aims to pull together both anchor and Echo representations in the unified latent space. 
% Note that 
Since this loss is applied in batches, % hence, 
it also pushes apart the rest of the representations in the batch, dividing the space and providing better representations. 
Finally, an additional uniformity term is computed to the anchor representation, to increase the tokens independency and enhance the overall contrastive objective: 

%Following U-MAE \cite{zhang_how_2022}, we additionally include a uniformity term to the anchor representation to complete the objective and provide it with dimensional collapse robustness. \ima{rephrase so we tell that we do it to further improve the contrastive as we encourage all dimensions to be used} Our final contrastive objective can be defined as follows:

\begin{equation*}
\mathcal{L}_{\text{CL}} =- \frac{1}{B} \sum_{i=1}^{B} \log \frac{ e^{z_i^T z'_i/\tau} }{ \sum_{j=1}^{B} e^{z_i^T z'_j/\tau} } + 0.1 \cdot \frac{1}{B} \sum_{i,j=1}^{B} \left(z_i \cdot z_j^T \right)^2
\end{equation*}
where $\mathbf{z}'$ and $\mathbf{z}$ represent the globally average pooled teacher and student outputs, respectively.

\subsection{Sorcen Unified Objective}

The final loss of Sorcen is a combination of Semantic reconstruction and Echo contrast objectives:
\begin{equation*}
L_{Sorcen} = L_{Recon} + \lambda \cdot L_{CL}
\end{equation*}
where $\lambda$ is the balancing hyperparameter of individual objectives. 
The final objective of Sorcen provides a unified representation learning objective carefully designed to improve both representation learning and image synthesis capacities of the model. Sorcen establishes as a new benchmark on unified SSL, proposing a novel symbiotic architecture and demonstrating strong SoTA performance. 

%Even if our method avoids the use of pixel augmentations, it is carefully designed and regularized to avoid collapses and instabilities. 
%Different to MAGE \cite{li_mage_2023}, our novel contrastive term manages to improve the quality of the representations for both generation and discrimination.  \ima{Improve closing}
\section{Results}
We conducted extensive experiments\footnote{Architecture and setup details are in the Section \ref{sec:setup} of the Appendix.} and analysis on generative and discriminative tasks, proving the capacity of Sorcen as a unified SSL framework. All experiments and pretraining (1600 epochs) are performed using ImageNet-1k \cite{krizhevsky_imagenet_2012}. For generation and discrimination evaluation, we followed % evaluation setup of 
MAGE \cite{li_mage_2023} and MaskGIT \cite{chang_maskgit_2022}, unless otherwise stated. For evaluation, only the Student encoder and the Decoder are used.

\subsection{General Evaluation}

\addtolength{\tabcolsep}{-0.075em}
\begin{table}[t]
\small
\begin{tabular}{llcclc}
\toprule
Methods                               & Model     & Acc. & FID   \\ \midrule
\textit{\textbf{Generative models}}   &           &      &       \\
%LDM-8    \cite{rombach_high-resolution_2022}                             &           & -    & 39.13 \\
DiT \cite{peebles_scalable_2023} (from \cite{chen_deconstructing_2024})                             & DiT       & 62.5 & 30.9  \\
%DiT-XL/2  \cite{peebles_scalable_2023} (from \cite{li_return_nodate})                            & DiT-XL    & -    & 27.32 \\
l-DAE          \cite{chen_deconstructing_2024}                       & DiT-L     & 69.6 & -     \\
Self-Cond. GAN  \cite{liu_diverse_2020}                &       -    & -    & 40.3 \\
BigGAN$_{256}$ \cite{brock_large_2018}                       &      -     & -    & 38.6  \\
%BigGAN$_{128}$  \cite{brock_large_2018}                         &      -     & -    & 23.56 \\
BigGAN+Clustering \cite{lucic_high-fidelity_2019}                     &     -      & -    & 22.0  \\
HiT        \cite{zhao_improved_2021}                           &       HiT    & -    & 30.8 \\
ADM           \cite{dhariwal_diffusion_2021}                        &      ADM     & -    & 26.2  \\
MaskGIT  \cite{chang_maskgit_2022}                             &  BERT         & 57.4    & 20.7 \\
BigBiGAN       \cite{donahue_large_2019}                       & RN50      & 56.6 & 21.6     \\
%MaskGIT     \cite{chang_maskgit_2022}                          & BERT      & 57.4 & -     \\
ViT-VQGAN      \cite{yu_vector-quantized_2021}                       & VIM-Base  & 65.1 & -     \\
ViT-VQGAN      \cite{yu_vector-quantized_2021}                       & VIM-Large & 73.2 & -     \\
IC-GAN        \cite{casanova_instance-conditioned_2021}                        &     -      & -    & 15.6  \\
ADDP           \cite{tian_addp_2024}                       & ViT-B     & 11.5 & 8.9   \\ \midrule
\textit{\textbf{Dual image crop models}} &           &      &       \\
MoCov3   \cite{chen_empirical_2021}                             & ViT-B     & 76.7 & -     \\
DINO  \cite{caron_emerging_2021}                                & ViT-B     & 72.8 & -     \\
iBot     \cite{zhou_ibot_2022}                             & ViT-B     & 76.0 & -     \\
MAGE-C    \cite{li_mage_2023}                            & ViT-B    & 78.2 & 31.8  \\
iBot$_{\textbf{12 crops}}$ \cite{zhou_ibot_2022}                       & ViT-B     & 79.5 & -     \\ \midrule
\midrule
\textit{\textbf{Single image crop models}} &           &      &       \\
MAE      \cite{he_masked_2022}                             & ViT-B     & 68.0 & -     \\
SimMIM   \cite{xie_simmim_2022}                             & ViT-B     & 57.8 & -     \\
U-MAE    \cite{zhang_how_2022}                             & ViT-L     & 65.8 & -     \\
MI-MAE   \cite{huang_learning_2024}                              & Swin-B     & 69.3 & -     \\
BeiT            \cite{bao_beit_2021}                      & ViT-B     & 56.7 & -     \\
CAE  \cite{chen_context_2024}                                & ViT-B     & 70.4 & -     \\
CMAE     \cite{huang_contrastive_2024}                               & ViT-B     & 73.9 & -     \\
XTRA     \cite{amrani_sample-_2024}                             & ViT-B     & 70.2 & -     \\
I-JEPA \cite{assran_self-supervised_2023}                                & ViT-B     & 72.9 & -     \\
MAGE      \cite{li_mage_2023}                            & ViT-B     & 74.7 & 11.1  \\
\rowcolor[HTML]{9AFF99} 
Sorcen                                & ViT-B     & \textbf{75.1} & \textbf{9.6}  \\ \bottomrule
\end{tabular}
\caption{\textbf{ImageNet-1k Evaluations on linear probing (Acc.) and unconditional image generation (FID).} We divide the comparison into Generative models, which mainly focus on image generation, Dual image crop models, Representation learning methods that leverage multiple augmented views that include extra information and Single image crop models, MIM and Contrastive methods that require a single image crop. Sorcen falls under this group.}\label{tab:hybrid}
\end{table}
\addtolength{\tabcolsep}{+0.075em}

\begin{figure}[t]%[htpb]
    \centering
    \includegraphics[width=\linewidth]{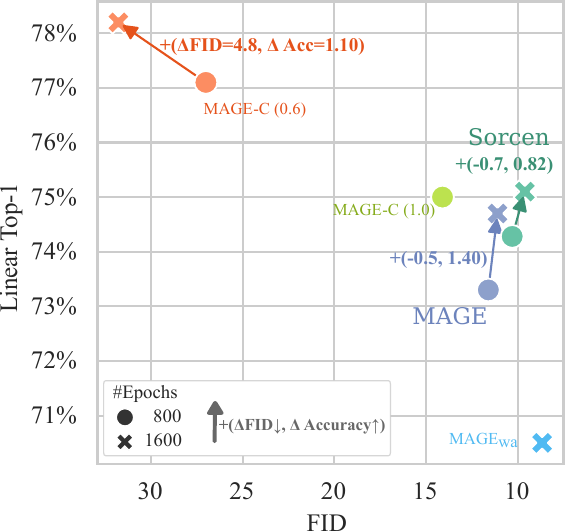}
    \caption{\textbf{Comparison of MAGE variants and Sorcen on IN1k.}  Arrows illustrate the performance shift from 800 to 1600 epochs pretraining. Sorcen pushes the Pareto front for unified methods in both pre-training durations. FID (X) axis is inverted.}
    \label{fig:magevssorcen}
\end{figure}

\noindent
\textbf{Unified Evaluation.} 
We compare the performance of Sorcen with other SoTA methods in Table \ref{tab:hybrid}.
%show the discriminative and generative performance of popular SoTA methods on linear probing and unconditional generation, respectively.
% For fair comparison, we only display methods trained and evaluated on IN1k, unless otherwise stated. 
We divide the comparison methods in three: 
(1) Generative methods: Focusing on %, whose priority falls on 
generation downstream tasks; 
(2) Dual image crop models: Trained with more than one view, implying multiple backward passes per training step and increased training information; 
(3) Single image crop models: Single image crop is used for every sample in the training set. 
Comparing to Generative models, Sorcen competes with recent methods such as ADDP \cite{tian_addp_2024}, while outperforming methods such as DiT \cite{peebles_scalable_2023}. 
Notably, none of the generative methods manage to outperform Sorcen in linear probing. 
For instance, ADDP \cite{tian_addp_2024} obtains good results on generation, however, dramatically underperforms on linear accuracy. 
ViT-VQGAN \cite{yu_vector-quantized_2021}, with its VIM-Large model (1697-M params) is a second-best competitor, % is the only one that competes, still under-performing by 
with a gap of 1.9\%. 
Qualitative image generation results are provided in Section \ref{qualitative} of Appendix.
% When compared with 2 
Comparing with Dual crop SSL methods, Sorcen outperforms DINO and competes with both MoCov3 \cite{chen_empirical_2021} and iBot \cite{zhou_image_2021}, which uses 2 augmentation views for every image in the dataset. 
Dual image strategy has increased the
% has proved to increase the 
stability and performance of iBot based models by $\sim$2\% \cite{moutakanni_you_2024} while adding significant overhead compared to single crop models.
% such as MAE \cite{he_masked_2022}. 
MoCov3 \cite{chen_empirical_2021}, a pure contrastive method, can't work without the augmentations. 
% Due to the characteristics of their systems, none of them provide 
None of the models provide any generative capacities and, consequently, they fall behind Sorcen in this aspect. 
MAGE-C \cite{li_mage_2023} on its own, sacrifices most of its generative capacities and increases its already huge VQGAN \cite{esser_taming_2021} overhead (2 passes are required now) to improve the linear accuracy, which \enquote{transforms} it into a discriminative only model. 
Sorcen clearly outperforms it on FID and competes in linear accuracy, while maintaining a good balance between discriminative and image synthesis capacities.
%staying loyal to hybrid line and showing a much better efficiency, 
%as explained in previous evaluations.
% Finally, we compare with models built over a single image view or 
With respect to single crop models, including vanilla MAGE \cite{li_mage_2023}, Sorcen outperforms previous SoTA, improving both linear accuracy and shows % providing 
superior generative capabilities. 
While we share some similarities with CAE \cite{chen_context_2024} and CMAE \cite{huang_contrastive_2024}, our novel contrastive approach, based on Echo samples prove to be superior to the pixel-based contrasts. % proposed by them. 
Sorcen provides a trailblazing single-crop contrastive strategy, achieving SoTA performance in unconditional image generation. 
Beyond outperforming single-crop methods, it also competes with multi-crop frameworks in image recognition.

\noindent
\textbf{MAGE vs Sorcen.}
Being the previous unified SSL SoTA, we compare Sorcen with all MAGE \cite{li_mage_2023} versions, including MAGE-C, its contrastive variant. 
As shown in Figure \ref{fig:magevssorcen}, we outperform vanilla MAGE under the same setup by 0.4\% on linear evaluation and 1.49 on FID. 
% When multiple MAGE \cite{li_mage_2023} versions are analysed, it can be seen how it fails to improve both metrics at the same time, being only able to increment one of them at the cost of the other. 
It can be noted that all MAGE versions fail to improve both metrics at the same time, being capable of incrementing only one % of them 
at the cost of the other.
MAGE-C, while superior in linear accuracy, is easily surpassed by Sorcen on FID metric\footnote{Table with numerical results in Section \ref{multi_setup} of the Appendix.}. 
On the other hand, MAGE$_{wa}$ falls behind Sorcen (4.6\% gap in linear) 
%increases the linear accuracy gap between Sorcen and MAGE to 4.6\% 
while marginally outperforming by 0.94 on FID. 
Different MAGE versions showcase various drawbacks that hinder the original idea of unified SSL. 
MAGE-C, for instance, requires a dual forward-backward pass (including the VQGAN \cite{esser_taming_2021}, see \textit{Efficiency comparison}) to perform a contrastive objective.
This however downgrades its capacity in generation. 
% As we can see in the scatter plot, 
The loss of generation capacities is more pronounced when the pretraining is longer, with a significant increase in FID (Figure \ref{fig:magevssorcen}). 
Inversely, MAGE$_{wa}$ dramatically decreases its linear performance to boost its generation. 
Sorcen, with a single crop setup, outperforms vanilla MAGE \cite{li_mage_2023} and competes with the \enquote{specialist} MAGE \cite{li_mage_2023} variants, proving itself as a novel and efficient unified SSL SoTA.
In terms of qualitative analysis, 
% we study in \Cref{sec:umaps} of the supplementary material the latent space of both MAGE and Sorcen, what supports 
we study the benefits of the our novel Echo contrast approach (\Cref{sec:umaps} of Appendix.) in improving discrimination without compromising and even improving reconstruction.

\noindent
\textbf{Few-shot Learning.} 
Following \cite{li_mage_2023, dosovitskiy_image_2021}, we evaluate % the capacity of 
Sorcen's capacity to produce useful features on low data regimes. 
% These shots 
We train a single linear layer for both models and evaluate on % before being evaluated on 
IN1k.
Compare with MAGE on 5, 10, 13 and 25 training shots of IN1k (Table \ref{tab:fewshot}), Sorcen consistently outperforms in all 4 setups with an average improvement of 1.76\%, showing a higher representation quality. % in low data regimes.

\begin{table}[t]
\centering
\small
\begin{tabular}{@{}lccccc@{}}
\toprule
\multirow{2}{*}{Method}               & \multicolumn{4}{c}{Shots per ImageNet Class} &       \\ \cmidrule(l){2-6} 
        & 5         & 10        & 13        & 25       & Avg.  \\ \midrule
MAGE & 48.44     & 57.37     & 59.84     & 63.66    & 57.33 \\
Sorcen        & \textbf{50.30}     & \textbf{59.21}     & \textbf{61.73}     & \textbf{65.13}    & \textbf{59.09} \\ \bottomrule
\end{tabular}
\caption{\textbf{Top-1 Accuracy for few-shot on IN1k for different number of shots per class}, as well as the average per method.}\label{tab:fewshot}
\end{table}

\noindent
\textbf{Transfer Learning.}
We explore the features generated by Sorcen by evaluating them on several transfer learning setups. 
Following the official data splits provided by \cite{zhou2022cocoop}, a total of 7 datasets are shown in Table \ref{tab:transfer}: Caltech101 \cite{fei2004learning}, Flowers102 \cite{nilsback2008automated}, OxfordPets \cite{parkhi2012cats}, SUN397 \cite{xiao2010sun}, DTD \cite{cimpoi2014describing}, UCF101 \cite{soomro2012ucf101}, and 
EuroSAT \cite{helber2019eurosat}. 
Sorcen outperforms MAGE in 5 out of 7 seven datasets with an average improvement of 1.53\%. In the remaining two datasets, it manages to obtain competitive results with less than 1\% gap. 
Our experiments\footnote{More transfer learning experiments in Section \ref{add_transfer} of Appendix.} prove that Sorcen, with its novel pretraining contrast, manages to learn better features that improve the generalization capacity of the backbone. 

\begin{figure}[t] %[htpb]
    \centering
    \includegraphics[width=\linewidth]{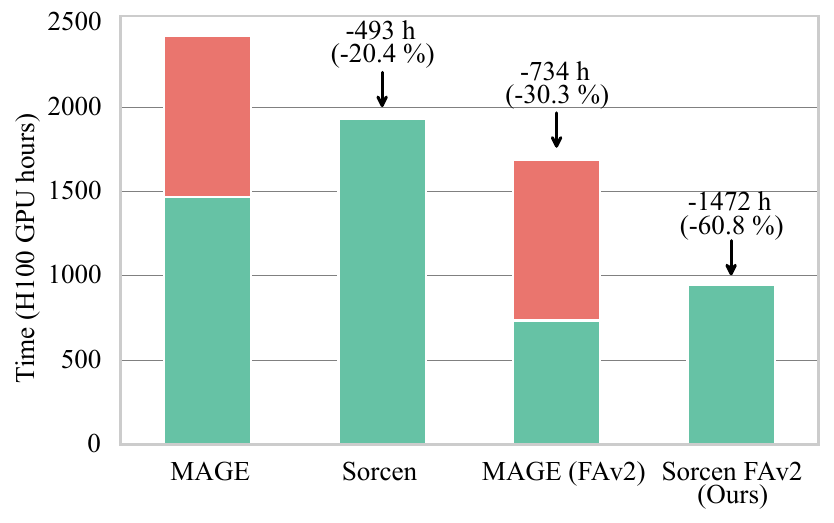}
    \caption{\textbf{Training time comparison for 1600 epochs of MAGE (left-most bar) and Sorcen (right-most bar). \textcolor[HTML]{EA756E}{$\blacksquare$} represents the VQGAN overhead.} FAv2 stands for FlashAttention v2 \cite{dao_flashattention-2_2023}.}
    \label{fig:barplot-times}
\end{figure}

\addtolength{\tabcolsep}{-0.4em}
\begin{figure*}
    \centering
    \small
    \begin{minipage}{0.6\textwidth}
        \centering
        \begin{tabular}{@{}lcccccccc@{}}
        \toprule
        Method        & Caltech & UCF101 & Flowers & Pets  & Sun   & EuroSAT & DTD   & Avg. \\ \midrule
        MAGE & 88.97   & 59.66  & \textbf{80.43}   & 71.38 & 52.36 & 64.44   & \textbf{53.90} & 67.31            \\
        Sorcen        & \textbf{89.61}   & \textbf{62.44}  & 79.50   & \textbf{71.93} & \textbf{53.26} & \textbf{71.28}   & 53.84 & \textbf{68.84}            \\ \bottomrule
        \end{tabular}
        \captionof{table}{\textbf{Transfer learning results (top-1 accuracy) for different datasets under 16-shot settings.} Last column contains the average across datasets.}\label{tab:transfer}
    \end{minipage}%
    \hfill
    \begin{minipage}{0.38\textwidth}
        \centering
        \small
        \begin{tabular}{@{}lccccc@{}}
        \toprule
        Method        & IN1k   & INv2    & Sketch & IN-R  & Avg. Drop \\ \midrule
        MAGE & 60.65 & 45.94 & 14.09  & 22.32 & 33.20        \\
        Sorcen        & \textbf{62.23} & \textbf{48.35} & \textbf{14.62}  & \textbf{25.27} & \textbf{32.82}        \\ \bottomrule
        \end{tabular}
        \captionof{table}{\textbf{K-NN top-1 accuracy in IN1k and 3 variants.} Average performance ($\downarrow$) drop in last column.}\label{tab:robustness}
    \end{minipage}
\end{figure*}
\addtolength{\tabcolsep}{+0.4em}

\noindent
\textbf{Model Robustness.}
Being able to produce robust representations is an important feature of any SSL algorithm. 
% In this evaluation, 
For this, we directly extract features of both Sorcen and MAGE and evaluate them against popular robustness and domain adaptation validation sets. 
% For the shake of 
To maintain the features of the encoder as clean as possible, we perform this evaluation using a simple k-NN over the globally pooled features of Sorcen and MAGE. 
As shown in Table \ref{tab:robustness}, Sorcen's features prove to be more accurate w.r.t IN1k \cite{krizhevsky_imagenet_2012} and shows a higher robustness when evaluated in the rest of the validation sets. 
Sorcen not only outperforms MAGE but also shows a smaller drop compared to its evaluation on IN1k.

\begin{table*}[t]
\centering
\small
\begin{tabular}{lccccccccc}
\toprule
                      & Echo & Echo Warm & JSM & Predictor & InfoNCE & Uniformity & L2 & Top-1          & FID                            \\ \midrule
Sorcen$_{NoEcho}$     &      &           & ✔   & ✔         & ✔       & ✔          &    & 68.06          & 23.82                          \\
Sorcen$_{NoWarmup}$   & ✔    &           & ✔   & ✔         & ✔       & ✔          &    & 33.20          & 47.17                          \\
Sorcen$_{NoJSM}$      & ✔    & ✔         &     & ✔         & ✔       & ✔          &    & 69.30          & 23.05                          \\
Sorcen$_{NoPred}$     & ✔    & ✔         & ✔   &           & ✔       & ✔          &    & 69.22          & \textbf{22.04}                 \\
Sorcen$_{Similarity}$ & ✔    & ✔         & ✔   & ✔         &         & ✔          & ✔  & 69.17          & 23.21                          \\
Sorcen$_{NoUniform}$  & ✔    & ✔         & ✔   & ✔         & ✔       &            &    & 69.12          & 23.66                              \\ \midrule
\rowcolor[HTML]{9AFF99} Sorcen                & ✔    & ✔         & ✔   & ✔         & ✔       &      ✔      &    & \textbf{69.62} & \underline{22.41} \\ \bottomrule
\end{tabular}
\caption{Ablation of different components of Sorcen. In each experiment (using IN200), an element of the proposal is removed during pre-training. The last two columns show results of each configuration in discriminative (few-shot top-1 accuracy) and generative (FID).}\label{tab:main_ablation}
\end{table*}

\noindent
\textbf{Efficiency Comparison.}
Previous unified methods relied heavily on VQGAN \cite{esser_taming_2021} inference during pretraining as they required common pixel augmentations such as Random Resized Crop. This introduces a significant overhead in those models. 
In contrast, % we propose Sorcen, which 
Sorcen eliminates the need for VQGAN \cite{esser_taming_2021} during pretraining by precomputing all tokens. % before training, 
% Sorcen achieves SoTA results in various tasks while significantly reducing training time. 
In Figure \ref{fig:barplot-times}, we compare the training times of MAGE \cite{li_mage_2023} and Sorcen. 
Both methods are executed under same conditions, 
% in the same setup, with training time measured in hours for 1,600 epochs 
on a single H100 GPU. 
We also report the comparisons with FlashAttentionv2 (FAv2) \cite{dao_flashattention-2_2023}, which is integrated into Sorcen's architecture. 
Sorcen is more efficient than MAGE \cite{li_mage_2023}, even after accounting for the additional teacher forward pass and JSM. Since our model relies solely on its transformer architecture, tools like FAv2 \cite{dao_flashattention-2_2023} prove more effective than in MAGE \cite{li_mage_2023}. 
% In fact, the hours 
The time required by Sorcen’s default setup with FAv2 \cite{dao_flashattention-2_2023} ($\sim$948 H100 hours) is almost identical to the overhead introduced by VQGAN \cite{esser_taming_2021} in MAGE ($\sim$951 H100 hours). 
Additionally, MAGE-C % the MAGE contrastive variant 
doubles this overhead due to the need for an extra view. 
Overall, Sorcen is more efficient, delivering superior results in a fraction of the training time required by MAGE \cite{li_mage_2023}. 
Further efficiency analysis in Sec. \ref{precomputed} and \ref{disk} of Appendix.

\subsection{Ablation Study}
% All the ablations in \Cref{tab:main_ablation} 
We highlight the importance of every design decision of Sorcen in \Cref{tab:main_ablation}. 
Due to computational limits, ablations were performed on IN200 (a 200-class subset of IN1k) for 200 epochs (1024 batch size), evaluated in a 25-shot regime.

%\addtolength{\tabcolsep}{-0.4em}
%\addtolength{\tabcolsep}{+0.4em}

\noindent
\textbf{(1) Echo Contrast.} Echo pair sampling and contrast effectively boosts both discriminative and generative performance.  \textit{Sorcen}$_{NoEcho}$ shows that Echo pairs improve Top-1 by +1.5\% (68.06\% vs 69.62\%) and reduce FID by 1.4 (23.82 vs 22.41). \textit{Sorcen}$_{NoWarmup}$ shows the need of a warmup (10 epochs in this case) to ensure that Echo samples are relevant for contrasting.
Without it, Top-1 drastically drops to 33.20\% and FID raises to 47.17.

%Our efficient echo pair sampling and contrast effectively boosts both discriminative and geneative performance. %When compared with $NoEcho$ version in Table \ref{tab:main_ablation}, where we train our model using the main input as positive pair, Echo pairs manage to improve the Top-1 accuracy by more than 1.5\%. 
%Comparing \textit{Sorcen}$_{NoEcho}$ with the full model in Table \ref{tab:main_ablation}, the introduction of echo pairs improves Top-1 accuracy by over 1.5\% (from 68.06\% to 69.62\%) and reduces the FID score by 1.4 points (from 23.82 to 22.41)
%Due to the roots of the echo samples, our system is not able to use them from the start and it requires a small warmup. As evidenced by \textit{Sorcen}$_{NoWarmup}$, removing the 10-epoch warmup phase leads to a substantial performance drop, with Top-1 accuracy plummeting to 33.20\% and FID soaring to 47.17. This highlights the necessity of the warmup for effective echo pair utilization.

\noindent\textbf{(2) JSM application.} %JSM, while not strictly required in this setup, further augments the benefits of Echo contrast. 
%The inclusion of JSM masking improves the Top-1 accuracy from 69.30\% in $NoJSM$ to 69.62\% in the full model.  Similarly, applying JSM contributes to the generation abilities of the model, with a contribution of more than 0.6 to FID reduction.
%We hypothesize that JSM's effectiveness is amplified with longer training and larger datasets. 
JSM further improves Echo contrast benefits. Adding JSM increases Top-1 from 69.30\% ($NoJSM$) to 69.62\% (full model) and reduces FID by 0.6.  We expect JSM to be more impactful with longer training and larger datasets. In such scenarios, stronger decoders benefit more from JSM's ability to diversify Echo pairs % by encouraging the model to capture 
capturing more nuanced semantic variations.

\noindent\textbf{(3) Predictors.}
Different to other Teacher-student architectures, Sorcen is robust without a predictor.  Removing it (\textit{Sorcen}$_{NoPred}$) only slightly decreases Top-1, % by 0.4\%, 
but improves FID, suggesting a trade-off: predictor favors discriminative, while its removal slightly favors generative quality. %Sorcen demonstrates robustness without relying on a predictor to prevent collapse.  Competitive results are achievable even without it. However, the predictor, in conjunction with the asymmetric architecture, provides a subtle yet beneficial boost to discriminative performance.: removing the predictor decreases the Top-1 accuracy by 0.4\%. On the other hand, the predictor hinders the generative capacities of the model with a slight worsening in terms of FID.
% This suggests a potential trade-off where the predictor prioritizes discriminative features, while its removal might slightly favor generative quality.

\noindent
\textbf{(4) InfoNCE vs Similarity Loss.} %Models based on reconstruction generate a very dense feature space that makes them less suitable for discriminative tasks. While designing Sorcen, we hypothesized that the capacity of InfoNCE loss to "push" negative samples would alleviate this problem and, consequently, be a much better alternative than similarity loss. While it is indeed a better option, we show in Table \ref{tab:main_ablation} that similarity loss can robustly provide competitive results in the tested setup (0.5\% difference in Top-1 accuracy). Possibly, the reduced batch size hindered the InfoNCEs capacity to push negative samples, providing less optimal results.
%Reconstruction-based models often produce dense feature spaces, potentially hindering discriminative tasks. We hypothesized that InfoNCE loss, with its negative sample "pushing" capability, would be more suitable than standard similarity loss for Sorcen.
%Our results confirm InfoNCE as the superior choice, although Similarity loss still yields competitive results: 0.5\% difference in Top-1 accuracy and 0.8 difference in FID.
%In terms of the image synthesis quality, Similarity loss yields a slightly higher FID score of 23.21 compared to 22.41 with InfoNCE, reinforcing the idea that InfoNCE contributes to a better hybrid feature space in our setup.
InfoNCE outperforms similarity loss, confirming our hypothesis that pushing negative samples is beneficial. While Similarity loss is competitive, % (0.5\% lower Top-1, 0.8 higher FID), 
InfoNCE yields a better unified feature space, though its full potential might be limited by batch size used. % in these ablations.
%These results prove that InfoNCE contributes to a better hybrid feature space in our setup. We posit that the limited batch size in our ablation studies might not fully unleash InfoNCE's capacity to leverage negative samples, potentially diminishing its advantage.

\noindent\textbf{(5) Uniformity Loss.} 
Uniformity loss encourages the feature distribution to be uniformly spread across the feature space, which can be beneficial for both representation learning and generation. 
\textit{Sorcen}$_{NoUniform}$ shows that removing it decreases Top-1 % (69.62\% to 69.12\%) 
and increases FID, % (22.41 to 23.66),
indicating a less smooth feature space and degraded generation without it. 
\subsection{Limitations and New Research Avenues}
While Sorcen has several strengths, it also has limitations to be explored in future research. 
%\textbf{Simple logit filtering and selection. }
(1) Although our simple logit filtering and selection process manages to produce rich echo samples, we hypothesize that more complex approaches, such as entropy-based sampling \cite{zhang_edt_2024}, common to NLP, could further improve the echo sample quality.
% of the Echo samples.
% On the other hand, 
(2) Sorcen is limited to single echo contrast, as multiple contrast would require more forward passes on the Teacher model. 
% We leave the exploration of an 
Efficient multi-echo contrast is a potential future research. %\textbf{Semantic token space on inference. }
(3) Even if pre-computing the training dataset removes all training overhead, the tokenization is still required for downstream tasks as most well-established evaluation pipelines depend on pixel augmentations.

\section{Conclusions}
In this work, we present Sorcen, a novel unified SSL framework, significantly outperforming previous SoTA methods in both representation learning and image synthesis.
Its synergic contrastive and reconstructive objectives % work synergistically, 
enhances both the model’s discrimination capabilities and its image generation performances. 
A key strength of Sorcen, the Echo Contrast, generates on-the-fly positive samples, enriching the contrastive objective with greater diversity without requiring external models or additional augmented crops. 
By operating on precomputed semantic tokens, Sorcen eliminates the need for token transformations during training, making it the most efficient unified SSL model within the SoTA. 
Extensive experiments and analysis on several discriminative tasks and unconditional image generation prove % that is possible to create a method that 
the capacity to not only excel in representation learning and image synthesis alone, but benefit from this duality. 
Moving forward we plan to extend Sorcen to other semantic token spaces, such as those provided by ViT-VQGAN, and to explore its potential on larger datasets.

% \newpage

% \input{sec/1_intro}
% \input{sec/2_formatting}
% \input{sec/3_finalcopy}
{
    \small
    \bibliographystyle{ieeenat_fullname}
    \bibliography{main}

\begin{thebibliography}{67}
\providecommand{\natexlab}[1]{#1}
\providecommand{\url}[1]{\texttt{#1}}
\expandafter\ifx\csname urlstyle\endcsname\relax
  \providecommand{\doi}[1]{doi: #1}\else
  \providecommand{\doi}{doi: \begingroup \urlstyle{rm}\Url}\fi

\bibitem[Amrani et~al.(2024)Amrani, Karlinsky, and Bronstein]{amrani_sample-_2024}
Elad Amrani, Leonid Karlinsky, and Alex Bronstein.
\newblock Sample- and {Parameter}-{Efficient} {Auto}-{Regressive} {Image} {Models}, 2024.
\newblock arXiv:2411.15648 [cs].

\bibitem[Assran et~al.(2023)Assran, Duval, Misra, Bojanowski, Vincent, Rabbat, LeCun, and Ballas]{assran_self-supervised_2023}
Mahmoud Assran, Quentin Duval, Ishan Misra, Piotr Bojanowski, Pascal Vincent, Michael Rabbat, Yann LeCun, and Nicolas Ballas.
\newblock Self-{Supervised} {Learning} {From} {Images} {With} a {Joint}-{Embedding} {Predictive} {Architecture}.
\newblock pages 15619--15629, 2023.

\bibitem[Bao et~al.(2021)Bao, Dong, Piao, and Wei]{bao_beit_2021}
Hangbo Bao, Li Dong, Songhao Piao, and Furu Wei.
\newblock {BEiT}: {BERT} {Pre}-{Training} of {Image} {Transformers}.
\newblock 2021.

\bibitem[Bender et~al.(2021)Bender, Gebru, McMillan-Major, and Shmitchell]{bender_dangers_2021}
Emily~M. Bender, Timnit Gebru, Angelina McMillan-Major, and Shmargaret Shmitchell.
\newblock On the {Dangers} of {Stochastic} {Parrots}: {Can} {Language} {Models} {Be} {Too} {Big}? ��.
\newblock In \emph{Proceedings of the 2021 {ACM} {Conference} on {Fairness}, {Accountability}, and {Transparency}}, pages 610--623, New York, NY, USA, 2021. Association for Computing Machinery.

\bibitem[Brock et~al.(2018)Brock, Donahue, and Simonyan]{brock_large_2018}
Andrew Brock, Jeff Donahue, and Karen Simonyan.
\newblock Large {Scale} {GAN} {Training} for {High} {Fidelity} {Natural} {Image} {Synthesis}.
\newblock 2018.

\bibitem[Caron et~al.(2018)Caron, Bojanowski, Joulin, and Douze]{caron_deep_2018}
Mathilde Caron, Piotr Bojanowski, Armand Joulin, and Matthijs Douze.
\newblock Deep {Clustering} for {Unsupervised} {Learning} of {Visual} {Features}.
\newblock pages 132--149, 2018.

\bibitem[Caron et~al.(2021)Caron, Touvron, Misra, Jégou, Mairal, Bojanowski, and Joulin]{caron_emerging_2021}
Mathilde Caron, Hugo Touvron, Ishan Misra, Hervé Jégou, Julien Mairal, Piotr Bojanowski, and Armand Joulin.
\newblock Emerging {Properties} in {Self}-{Supervised} {Vision} {Transformers}.
\newblock pages 9650--9660, 2021.

\bibitem[Casanova et~al.(2021)Casanova, Careil, Verbeek, Drozdzal, and Romero~Soriano]{casanova_instance-conditioned_2021}
Arantxa Casanova, Marlene Careil, Jakob Verbeek, Michal Drozdzal, and Adriana Romero~Soriano.
\newblock Instance-{Conditioned} {GAN}.
\newblock In \emph{Advances in {Neural} {Information} {Processing} {Systems}}, pages 27517--27529. Curran Associates, Inc., 2021.

\bibitem[Chang et~al.(2022)Chang, Zhang, Jiang, Liu, and Freeman]{chang_maskgit_2022}
Huiwen Chang, Han Zhang, Lu Jiang, Ce Liu, and William~T. Freeman.
\newblock {MaskGIT}: {Masked} {Generative} {Image} {Transformer}.
\newblock pages 11315--11325, 2022.

\bibitem[Chen et~al.(2020{\natexlab{a}})Chen, Kornblith, Norouzi, and Hinton]{chen_simple_2020}
Ting Chen, Simon Kornblith, Mohammad Norouzi, and Geoffrey Hinton.
\newblock A {Simple} {Framework} for {Contrastive} {Learning} of {Visual} {Representations}.
\newblock In \emph{Proceedings of the 37th {International} {Conference} on {Machine} {Learning}}, pages 1597--1607. PMLR, 2020{\natexlab{a}}.
\newblock ISSN: 2640-3498.

\bibitem[Chen and He(2021)]{chen_exploring_2021}
Xinlei Chen and Kaiming He.
\newblock Exploring {Simple} {Siamese} {Representation} {Learning}.
\newblock In \emph{2021 {IEEE}/{CVF} {Conference} on {Computer} {Vision} and {Pattern} {Recognition} ({CVPR})}, pages 15745--15753, Nashville, TN, USA, 2021. IEEE.

\bibitem[Chen et~al.(2020{\natexlab{b}})Chen, Fan, Girshick, and He]{chen_improved_2020}
Xinlei Chen, Haoqi Fan, Ross Girshick, and Kaiming He.
\newblock Improved {Baselines} with {Momentum} {Contrastive} {Learning}, 2020{\natexlab{b}}.
\newblock arXiv:2003.04297 [cs].

\bibitem[Chen et~al.(2021)Chen, Xie, and He]{chen_empirical_2021}
Xinlei Chen, Saining Xie, and Kaiming He.
\newblock An {Empirical} {Study} of {Training} {Self}-{Supervised} {Vision} {Transformers}.
\newblock pages 9640--9649, 2021.

\bibitem[Chen et~al.(2024{\natexlab{a}})Chen, Ding, Wang, Xin, Mo, Wang, Han, Luo, Zeng, and Wang]{chen_context_2024}
Xiaokang Chen, Mingyu Ding, Xiaodi Wang, Ying Xin, Shentong Mo, Yunhao Wang, Shumin Han, Ping Luo, Gang Zeng, and Jingdong Wang.
\newblock Context {Autoencoder} for {Self}-supervised {Representation} {Learning}.
\newblock \emph{International Journal of Computer Vision}, 132\penalty0 (1):\penalty0 208--223, 2024{\natexlab{a}}.

\bibitem[Chen et~al.(2024{\natexlab{b}})Chen, Liu, Xie, and He]{chen_deconstructing_2024}
Xinlei Chen, Zhuang Liu, Saining Xie, and Kaiming He.
\newblock Deconstructing {Denoising} {Diffusion} {Models} for {Self}-{Supervised} {Learning}, 2024{\natexlab{b}}.
\newblock arXiv:2401.14404 [cs].

\bibitem[Chen et~al.(2024{\natexlab{c}})Chen, Hu, Chen, and Zhang]{chen_progress_2024}
Zhihua Chen, Bo Hu, Zhongsheng Chen, and Jiarui Zhang.
\newblock Progress and {Thinking} on {Self}-{Supervised} {Learning} {Methods} in {Computer} {Vision}: {A} {Review}.
\newblock \emph{IEEE Sensors Journal}, 24\penalty0 (19):\penalty0 29524--29544, 2024{\natexlab{c}}.
\newblock Conference Name: IEEE Sensors Journal.

\bibitem[Cimpoi et~al.(2014)Cimpoi, Maji, Kokkinos, Mohamed, and Vedaldi]{cimpoi2014describing}
Mircea Cimpoi, Subhransu Maji, Iasonas Kokkinos, Sammy Mohamed, and Andrea Vedaldi.
\newblock Describing textures in the wild.
\newblock In \emph{Proceedings of the IEEE conference on computer vision and pattern recognition}, pages 3606--3613, 2014.

\bibitem[Dao(2023)]{dao_flashattention-2_2023}
Tri Dao.
\newblock {FlashAttention}-2: {Faster} {Attention} with {Better} {Parallelism} and {Work} {Partitioning}.
\newblock 2023.

\bibitem[Dhariwal and Nichol(2021)]{dhariwal_diffusion_2021}
Prafulla Dhariwal and Alex Nichol.
\newblock Diffusion models beat {GANs} on image synthesis.
\newblock In \emph{Proceedings of the 35th {International} {Conference} on {Neural} {Information} {Processing} {Systems}}, pages 8780--8794, Red Hook, NY, USA, 2021. Curran Associates Inc.

\bibitem[Donahue and Simonyan(2019)]{donahue_large_2019}
Jeff Donahue and Karen Simonyan.
\newblock Large {Scale} {Adversarial} {Representation} {Learning}.
\newblock In \emph{Advances in {Neural} {Information} {Processing} {Systems}}. Curran Associates, Inc., 2019.

\bibitem[Dosovitskiy et~al.(2021)Dosovitskiy, Beyer, Kolesnikov, Weissenborn, Zhai, Unterthiner, Dehghani, Minderer, Heigold, Gelly, Uszkoreit, and Houlsby]{dosovitskiy_image_2021}
Alexey Dosovitskiy, Lucas Beyer, Alexander Kolesnikov, Dirk Weissenborn, Xiaohua Zhai, Thomas Unterthiner, Mostafa Dehghani, Matthias Minderer, Georg Heigold, Sylvain Gelly, Jakob Uszkoreit, and Neil Houlsby.
\newblock An {Image} is {Worth} 16x16 {Words}: {Transformers} for {Image} {Recognition} at {Scale}, 2021.
\newblock arXiv:2010.11929 [cs].

\bibitem[Dwibedi et~al.(2021)Dwibedi, Aytar, Tompson, Sermanet, and Zisserman]{dwibedi_little_2021}
Debidatta Dwibedi, Yusuf Aytar, Jonathan Tompson, Pierre Sermanet, and Andrew Zisserman.
\newblock With a {Little} {Help} {From} {My} {Friends}: {Nearest}-{Neighbor} {Contrastive} {Learning} of {Visual} {Representations}.
\newblock pages 9588--9597, 2021.

\bibitem[Esser et~al.(2021)Esser, Rombach, and Ommer]{esser_taming_2021}
Patrick Esser, Robin Rombach, and Bjorn Ommer.
\newblock Taming {Transformers} for {High}-{Resolution} {Image} {Synthesis}.
\newblock pages 12873--12883, 2021.

\bibitem[Estepa et~al.(2023{\natexlab{a}})Estepa, Rodr{\'\i}guez-de Vera, Nagarajan, and Radeva]{estepa2023good}
Imanol~G Estepa, Jes{\'u}s Rodr{\'\i}guez-de Vera, Bhalaji Nagarajan, and Petia Radeva.
\newblock Good fences make good neighbours.
\newblock In \emph{Proceedings of the IEEE/CVF International Conference on Computer Vision}, pages 216--226, 2023{\natexlab{a}}.

\bibitem[Estepa et~al.(2023{\natexlab{b}})Estepa, Sarasua, Nagarajan, and Radeva]{estepa_all4one_2023}
Imanol~G. Estepa, Ignacio Sarasua, Bhalaji Nagarajan, and Petia Radeva.
\newblock {All4One}: {Symbiotic} {Neighbour} {Contrastive} {Learning} via {Self}-{Attention} and {Redundancy} {Reduction}.
\newblock pages 16243--16253, 2023{\natexlab{b}}.

\bibitem[Fei-Fei et~al.(2004)Fei-Fei, Fergus, and Perona]{fei2004learning}
Li Fei-Fei, Rob Fergus, and Pietro Perona.
\newblock Learning generative visual models from few training examples: An incremental bayesian approach tested on 101 object categories.
\newblock In \emph{2004 conference on computer vision and pattern recognition workshop}, pages 178--178. IEEE, 2004.

\bibitem[Grill et~al.(2020)Grill, Strub, Altché, Tallec, Richemond, Buchatskaya, Doersch, Avila~Pires, Guo, Gheshlaghi~Azar, Piot, kavukcuoglu, Munos, and Valko]{grill_bootstrap_2020}
Jean-Bastien Grill, Florian Strub, Florent Altché, Corentin Tallec, Pierre Richemond, Elena Buchatskaya, Carl Doersch, Bernardo Avila~Pires, Zhaohan Guo, Mohammad Gheshlaghi~Azar, Bilal Piot, koray kavukcuoglu, Remi Munos, and Michal Valko.
\newblock Bootstrap {Your} {Own} {Latent} - {A} {New} {Approach} to {Self}-{Supervised} {Learning}.
\newblock In \emph{Advances in {Neural} {Information} {Processing} {Systems}}, pages 21271--21284. Curran Associates, Inc., 2020.

\bibitem[He et~al.(2022)He, Chen, Xie, Li, Dollár, and Girshick]{he_masked_2022}
Kaiming He, Xinlei Chen, Saining Xie, Yanghao Li, Piotr Dollár, and Ross Girshick.
\newblock Masked {Autoencoders} {Are} {Scalable} {Vision} {Learners}.
\newblock pages 16000--16009, 2022.

\bibitem[Helber et~al.(2019)Helber, Bischke, Dengel, and Borth]{helber2019eurosat}
Patrick Helber, Benjamin Bischke, Andreas Dengel, and Damian Borth.
\newblock Eurosat: A novel dataset and deep learning benchmark for land use and land cover classification.
\newblock \emph{IEEE Journal of Selected Topics in Applied Earth Observations and Remote Sensing}, 12\penalty0 (7):\penalty0 2217--2226, 2019.

\bibitem[Hinojosa et~al.(2025)Hinojosa, Liu, and Ghanem]{hinojosa_colormae_2025}
Carlos Hinojosa, Shuming Liu, and Bernard Ghanem.
\newblock {ColorMAE}: {Exploring} {Data}-{Independent} {Masking} {Strategies} in {Masked} {AutoEncoders}.
\newblock In \emph{Computer {Vision} – {ECCV} 2024}, pages 432--449, Cham, 2025. Springer Nature Switzerland.

\bibitem[Hondru et~al.(2025)Hondru, Croitoru, Minaee, Ionescu, and Sebe]{hondru_masked_2025}
Vlad Hondru, Florinel~Alin Croitoru, Shervin Minaee, Radu~Tudor Ionescu, and Nicu Sebe.
\newblock Masked {Image} {Modeling}: {A} {Survey}, 2025.
\newblock arXiv:2408.06687 [cs].

\bibitem[Hu et~al.(2023)Hu, Chen, Wang, Li, Wang, Sun, and Li]{hu_complexity_2023}
Tianyang Hu, Fei Chen, Haonan Wang, Jiawei Li, Wenjia Wang, Jiacheng Sun, and Zhenguo Li.
\newblock Complexity {Matters}: {Rethinking} the {Latent} {Space} for {Generative} {Modeling}, 2023.
\newblock arXiv:2307.08283 [cs].

\bibitem[Huang et~al.(2024{\natexlab{a}})Huang, Ma, You, and Xu]{huang_learning_2024}
Tao Huang, Yanxiang Ma, Shan You, and Chang Xu.
\newblock Learning {Mask} {Invariant} {Mutual} {Information} for {Masked} {Image} {Modeling}.
\newblock 2024{\natexlab{a}}.

\bibitem[Huang et~al.(2024{\natexlab{b}})Huang, Jin, Lu, Hou, Cheng, Fu, Shen, and Feng]{huang_contrastive_2024}
Zhicheng Huang, Xiaojie Jin, Chengze Lu, Qibin Hou, Ming-Ming Cheng, Dongmei Fu, Xiaohui Shen, and Jiashi Feng.
\newblock Contrastive {Masked} {Autoencoders} are {Stronger} {Vision} {Learners}.
\newblock \emph{IEEE Transactions on Pattern Analysis and Machine Intelligence}, 46\penalty0 (4):\penalty0 2506--2517, 2024{\natexlab{b}}.
\newblock Conference Name: IEEE Transactions on Pattern Analysis and Machine Intelligence.

\bibitem[Jiang et~al.(2021)Jiang, Zhang, Talwar, and Mozer]{jiang_characterizing_2021}
Ziheng Jiang, Chiyuan Zhang, Kunal Talwar, and Michael~C. Mozer.
\newblock Characterizing {Structural} {Regularities} of {Labeled} {Data} in {Overparameterized} {Models}.
\newblock In \emph{Proceedings of the 38th {International} {Conference} on {Machine} {Learning}}, pages 5034--5044. PMLR, 2021.
\newblock ISSN: 2640-3498.

\bibitem[Krizhevsky et~al.(2012)Krizhevsky, Sutskever, and Hinton]{krizhevsky_imagenet_2012}
Alex Krizhevsky, Ilya Sutskever, and Geoffrey~E Hinton.
\newblock {ImageNet} {Classification} with {Deep} {Convolutional} {Neural} {Networks}.
\newblock In \emph{Advances in {Neural} {Information} {Processing} {Systems}}. Curran Associates, Inc., 2012.

\bibitem[Lee et~al.(2025)Lee, Park, Heo, Han, and Shim]{lee_seit_2025}
Minhyun Lee, Song Park, Byeongho Heo, Dongyoon Han, and Hyunjung Shim.
\newblock {SeiT}++: {Masked} {Token} {Modeling} {Improves} {Storage}-{Efficient} {Training}.
\newblock In \emph{Computer {Vision} – {ECCV} 2024}, pages 180--197, Cham, 2025. Springer Nature Switzerland.

\bibitem[Li et~al.(2023)Li, Chang, Mishra, Zhang, Katabi, and Krishnan]{li_mage_2023}
Tianhong Li, Huiwen Chang, Shlok Mishra, Han Zhang, Dina Katabi, and Dilip Krishnan.
\newblock {MAGE}: {MAsked} {Generative} {Encoder} {To} {Unify} {Representation} {Learning} and {Image} {Synthesis}.
\newblock pages 2142--2152, 2023.

\bibitem[Liu et~al.(2020)Liu, Wang, Bau, Zhu, and Torralba]{liu_diverse_2020}
Steven Liu, Tongzhou Wang, David Bau, Jun-Yan Zhu, and Antonio Torralba.
\newblock Diverse {Image} {Generation} via {Self}-{Conditioned} {GANs}.
\newblock pages 14286--14295, 2020.

\bibitem[Lučić et~al.(2019)Lučić, Tschannen, Ritter, Zhai, Bachem, and Gelly]{lucic_high-fidelity_2019}
Mario Lučić, Michael Tschannen, Marvin Ritter, Xiaohua Zhai, Olivier Bachem, and Sylvain Gelly.
\newblock High-{Fidelity} {Image} {Generation} {With} {Fewer} {Labels}.
\newblock In \emph{Proceedings of the 36th {International} {Conference} on {Machine} {Learning}}, pages 4183--4192. PMLR, 2019.
\newblock ISSN: 2640-3498.

\bibitem[McInnes et~al.(2018)McInnes, Healy, and Melville]{mcinnes2018umap}
Leland McInnes, John Healy, and James Melville.
\newblock Umap: Uniform manifold approximation and projection for dimension reduction.
\newblock \emph{arXiv preprint arXiv:1802.03426}, 2018.

\bibitem[Moutakanni et~al.(2024)Moutakanni, Oquab, Szafraniec, Vakalopoulou, and Bojanowski]{moutakanni_you_2024}
Théo Moutakanni, Maxime Oquab, Marc Szafraniec, Maria Vakalopoulou, and Piotr Bojanowski.
\newblock You {Don}'t {Need} {Domain}-{Specific} {Data} {Augmentations} {When} {Scaling} {Self}-{Supervised} {Learning}, 2024.
\newblock arXiv:2406.09294 [cs].

\bibitem[Nilsback and Zisserman(2008)]{nilsback2008automated}
Maria-Elena Nilsback and Andrew Zisserman.
\newblock Automated flower classification over a large number of classes.
\newblock In \emph{2008 Sixth Indian conference on computer vision, graphics \& image processing}, pages 722--729. IEEE, 2008.

\bibitem[Oord et~al.(2019)Oord, Li, and Vinyals]{oord_representation_2019}
Aaron van~den Oord, Yazhe Li, and Oriol Vinyals.
\newblock Representation {Learning} with {Contrastive} {Predictive} {Coding}, 2019.
\newblock arXiv:1807.03748 [cs, stat].

\bibitem[Oquab et~al.(2023)Oquab, Darcet, Moutakanni, Vo, Szafraniec, Khalidov, Fernandez, Haziza, Massa, El-Nouby, Assran, Ballas, Galuba, Howes, Huang, Li, Misra, Rabbat, Sharma, Synnaeve, Xu, Jegou, Mairal, Labatut, Joulin, and Bojanowski]{oquab_dinov2_2023}
Maxime Oquab, Timothée Darcet, Théo Moutakanni, Huy~V. Vo, Marc Szafraniec, Vasil Khalidov, Pierre Fernandez, Daniel Haziza, Francisco Massa, Alaaeldin El-Nouby, Mido Assran, Nicolas Ballas, Wojciech Galuba, Russell Howes, Po-Yao Huang, Shang-Wen Li, Ishan Misra, Michael Rabbat, Vasu Sharma, Gabriel Synnaeve, Hu Xu, Herve Jegou, Julien Mairal, Patrick Labatut, Armand Joulin, and Piotr Bojanowski.
\newblock {DINOv2}: {Learning} {Robust} {Visual} {Features} without {Supervision}.
\newblock \emph{Transactions on Machine Learning Research}, 2023.

\bibitem[Ozbulak et~al.(2023)Ozbulak, Lee, Boga, Anzaku, Park, Messem, Neve, and Vankerschaver]{ozbulak_know_2023}
Utku Ozbulak, Hyun~Jung Lee, Beril Boga, Esla~Timothy Anzaku, Ho-min Park, Arnout~Van Messem, Wesley~De Neve, and Joris Vankerschaver.
\newblock Know {Your} {Self}-supervised {Learning}: {A} {Survey} on {Image}-based {Generative} and {Discriminative} {Training}.
\newblock \emph{Transactions on Machine Learning Research}, 2023.

\bibitem[Park et~al.(2023)Park, Chun, Heo, Kim, and Yun]{park_seit_2023}
Song Park, Sanghyuk Chun, Byeongho Heo, Wonjae Kim, and Sangdoo Yun.
\newblock {SeiT}: {Storage}-{Efficient} {Vision} {Training} with {Tokens} {Using} 1\% of {Pixel} {Storage}.
\newblock pages 17248--17259, 2023.

\bibitem[Parkhi et~al.(2012)Parkhi, Vedaldi, Zisserman, and Jawahar]{parkhi2012cats}
Omkar~M Parkhi, Andrea Vedaldi, Andrew Zisserman, and CV Jawahar.
\newblock Cats and dogs.
\newblock In \emph{2012 IEEE conference on computer vision and pattern recognition}, pages 3498--3505. IEEE, 2012.

\bibitem[Peebles and Xie(2023)]{peebles_scalable_2023}
William Peebles and Saining Xie.
\newblock Scalable {Diffusion} {Models} with {Transformers}.
\newblock pages 4195--4205, 2023.

\bibitem[Rombach et~al.(2022)Rombach, Blattmann, Lorenz, Esser, and Ommer]{rombach_high-resolution_2022}
Robin Rombach, Andreas Blattmann, Dominik Lorenz, Patrick Esser, and Björn Ommer.
\newblock High-{Resolution} {Image} {Synthesis} {With} {Latent} {Diffusion} {Models}.
\newblock pages 10684--10695, 2022.

\bibitem[Soomro et~al.(2012)Soomro, Zamir, and Shah]{soomro2012ucf101}
Khurram Soomro, Amir~Roshan Zamir, and Mubarak Shah.
\newblock Ucf101: A dataset of 101 human actions classes from videos in the wild.
\newblock \emph{arXiv preprint arXiv:1212.0402}, 2012.

\bibitem[Tian et~al.(2024)Tian, Tao, Dai, Li, Li, Lu, Wang, Li, Huang, and Zhu]{tian_addp_2024}
Changyao Tian, Chenxin Tao, Jifeng Dai, Hao Li, Ziheng Li, Lewei Lu, Xiaogang Wang, Hongsheng Li, Gao Huang, and Xizhou Zhu.
\newblock {ADDP}: {Learning} {General} {Representations} for {Image} {Recognition} and {Generation} with {Alternating} {Denoising} {Diffusion} {Process}, 2024.
\newblock arXiv:2306.05423 [cs].

\bibitem[Vaswani et~al.(2017)Vaswani, Shazeer, Parmar, Uszkoreit, Jones, Gomez, Kaiser, and Polosukhin]{vaswani_attention_2017}
Ashish Vaswani, Noam Shazeer, Niki Parmar, Jakob Uszkoreit, Llion Jones, Aidan~N Gomez, Łukasz Kaiser, and Illia Polosukhin.
\newblock Attention is {All} you {Need}.
\newblock In \emph{Advances in {Neural} {Information} {Processing} {Systems}}. Curran Associates, Inc., 2017.

\bibitem[Xiao et~al.(2010)Xiao, Hays, Ehinger, Oliva, and Torralba]{xiao2010sun}
Jianxiong Xiao, James Hays, Krista~A Ehinger, Aude Oliva, and Antonio Torralba.
\newblock Sun database: Large-scale scene recognition from abbey to zoo.
\newblock In \emph{2010 IEEE computer society conference on computer vision and pattern recognition}, pages 3485--3492. IEEE, 2010.

\bibitem[Xie et~al.(2022)Xie, Zhang, Cao, Lin, Bao, Yao, Dai, and Hu]{xie_simmim_2022}
Zhenda Xie, Zheng Zhang, Yue Cao, Yutong Lin, Jianmin Bao, Zhuliang Yao, Qi Dai, and Han Hu.
\newblock {SimMIM}: {A} {Simple} {Framework} for {Masked} {Image} {Modeling}.
\newblock pages 9653--9663, 2022.

\bibitem[Yang et~al.(2023)Yang, Zhang, Song, Hong, Xu, Zhao, Zhang, Cui, and Yang]{yang_diffusion_2023}
Ling Yang, Zhilong Zhang, Yang Song, Shenda Hong, Runsheng Xu, Yue Zhao, Wentao Zhang, Bin Cui, and Ming-Hsuan Yang.
\newblock Diffusion {Models}: {A} {Comprehensive} {Survey} of {Methods} and {Applications}.
\newblock \emph{ACM Comput. Surv.}, 56\penalty0 (4):\penalty0 105:1--105:39, 2023.

\bibitem[Yu et~al.(2021)Yu, Li, Koh, Zhang, Pang, Qin, Ku, Xu, Baldridge, and Wu]{yu_vector-quantized_2021}
Jiahui Yu, Xin Li, Jing~Yu Koh, Han Zhang, Ruoming Pang, James Qin, Alexander Ku, Yuanzhong Xu, Jason Baldridge, and Yonghui Wu.
\newblock Vector-quantized {Image} {Modeling} with {Improved} {VQGAN}.
\newblock 2021.

\bibitem[Zbontar et~al.(2021)Zbontar, Jing, Misra, LeCun, and Deny]{zbontar_barlow_2021}
Jure Zbontar, Li Jing, Ishan Misra, Yann LeCun, and Stephane Deny.
\newblock Barlow {Twins}: {Self}-{Supervised} {Learning} via {Redundancy} {Reduction}.
\newblock In \emph{Proceedings of the 38th {International} {Conference} on {Machine} {Learning}}, pages 12310--12320. PMLR, 2021.
\newblock ISSN: 2640-3498.

\bibitem[Zeng et~al.(2025)Zeng, Wu, Hu, Xu, and Shi]{zeng_contrastive_2025}
Dewen Zeng, Yawen Wu, Xinrong Hu, Xiaowei Xu, and Yiyu Shi.
\newblock Contrastive {Learning} with {Synthetic} {Positives}.
\newblock In \emph{Computer {Vision} – {ECCV} 2024}, pages 430--447, Cham, 2025. Springer Nature Switzerland.

\bibitem[Zhang et~al.(2022)Zhang, Wang, and Wang]{zhang_how_2022}
Qi Zhang, Yifei Wang, and Yisen Wang.
\newblock How {Mask} {Matters}: {Towards} {Theoretical} {Understandings} of {Masked} {Autoencoders}.
\newblock \emph{Advances in Neural Information Processing Systems}, 35:\penalty0 27127--27139, 2022.

\bibitem[Zhang et~al.(2024{\natexlab{a}})Zhang, Bao, and Huang]{zhang_edt_2024}
Shimao Zhang, Yu Bao, and Shujian Huang.
\newblock {EDT}: {Improving} {Large} {Language} {Models}' {Generation} by {Entropy}-based {Dynamic} {Temperature} {Sampling}, 2024{\natexlab{a}}.
\newblock arXiv:2403.14541 [cs].

\bibitem[Zhang et~al.(2024{\natexlab{b}})Zhang, Zhang, Lin, Xing, Mo, Huang, Xie, Li, Luan, Zhao, Zhang, and Chen]{zhang_towards_2024}
Zhanjie Zhang, Quanwei Zhang, Huaizhong Lin, Wei Xing, Juncheng Mo, Shuaicheng Huang, Jinheng Xie, Guangyuan Li, Junsheng Luan, Lei Zhao, Dalong Zhang, and Lixia Chen.
\newblock Towards highly realistic artistic style transfer via stable diffusion with step-aware and layer-aware prompt.
\newblock In \emph{Proceedings of the {Thirty}-{Third} {International} {Joint} {Conference} on {Artificial} {Intelligence}}, pages 7814--7822, Jeju, Korea, 2024{\natexlab{b}}.

\bibitem[Zhao et~al.(2021)Zhao, Zhang, Chen, Metaxas, and Zhang]{zhao_improved_2021}
Long Zhao, Zizhao Zhang, Ting Chen, Dimitris Metaxas, and Han Zhang.
\newblock Improved {Transformer} for {High}-{Resolution} {GANs}.
\newblock In \emph{Advances in {Neural} {Information} {Processing} {Systems}}, pages 18367--18380. Curran Associates, Inc., 2021.

\bibitem[Zhou et~al.(2021)Zhou, Wei, Wang, Shen, Xie, Yuille, and Kong]{zhou_image_2021}
Jinghao Zhou, Chen Wei, Huiyu Wang, Wei Shen, Cihang Xie, Alan Yuille, and Tao Kong.
\newblock Image {BERT} {Pre}-training with {Online} {Tokenizer}.
\newblock 2021.

\bibitem[Zhou et~al.(2022{\natexlab{a}})Zhou, Wei, Wang, Shen, Xie, Yuille, and Kong]{zhou_ibot_2022}
Jinghao Zhou, Chen Wei, Huiyu Wang, Wei Shen, Cihang Xie, Alan Yuille, and Tao Kong.
\newblock {iBOT}: {Image} {BERT} {Pre}-{Training} with {Online} {Tokenizer}, 2022{\natexlab{a}}.
\newblock arXiv:2111.07832 [cs].

\bibitem[Zhou et~al.(2022{\natexlab{b}})Zhou, Yang, Loy, and Liu]{zhou2022cocoop}
Kaiyang Zhou, Jingkang Yang, Chen~Change Loy, and Ziwei Liu.
\newblock Conditional prompt learning for vision-language models.
\newblock In \emph{IEEE/CVF Conference on Computer Vision and Pattern Recognition (CVPR)}, 2022{\natexlab{b}}.

\bibitem[Zhu et~al.(2024)Zhu, Li, Zhang, Li, Xu, and Bing]{zhu_stabilize_2024}
Yongxin Zhu, Bocheng Li, Hang Zhang, Xin Li, Linli Xu, and Lidong Bing.
\newblock Stabilize the {Latent} {Space} for {Image} {Autoregressive} {Modeling}: {A} {Unified} {Perspective}, 2024.
\newblock arXiv:2410.12490 [cs].

\end{thebibliography}
}

\clearpage
\setcounter{page}{1}
\maketitlesupplementary
\renewcommand{\thesection}{\Alph{section}}
\renewcommand{\theHsection}{\Alph{section}}

\setcounter{section}{0}

\section{Implementation Details}\label{sec:setup}

\textbf{Architecture details.} 
% Sorcen divides its contrastive pretraining into two different branches, teacher and student branch. 
Before the contrastive loss is applied in Sorcen, the output of the student branch is passed through a simple MLP projector of 2 layers and an MLP predictor of another 2 layers. 
The teacher branch is composed of a smoothed version of the student branch, including the encoder and the projector which are updated via EMA every training step. 
The teacher branch does not include any predictor. 
Reconstruction pretraining exclusively follows the student branch and includes an additional Decoder to perform the semantic reconstruction. For all experiments we use $\lambda=0.1$, $K=15$ and a projector size of 512 according to analysis provided in \Cref{sec:hyperparameters-values}.

\noindent
\textbf{Evaluation setup.} 
We apply 20 steps to generate images for generative evaluation and all discriminative tasks are conducted by globally averaging the features generated by the student encoder. 
Teacher encoder, projectors and predictor are completely discarded on inference, using the Student encoder for discriminative tasks and the decoder for generative tasks,  unless otherwise stated. To ensure reproducibility and provide a comprehensive understanding of our experimental setup, we detail the implementation specifics in this section. Tables \ref{tab:pretraining_settings}, \ref{tab:linear_probing_settings}, and \ref{tab:few_shot_settings} summarize the key hyperparameter settings used for pre-training, linear probing, and few-shot learning experiments, respectively.  %These tables specify crucial configurations including optimizers, learning rates, batch sizes, and other relevant training parameters.

\begin{table}[!htpb]
\centering
\begin{tabular}{|c|c|}
\hline
\textbf{config} & \textbf{value} \\
\hline
optimizer & AdamW  \\
base learning rate & 1.5e-4 \\
weight decay & 0.05 \\
optimizer momentum & $\beta_1$, $\beta_2$ = 0.9, 0.95 \\
batch size & 4096\\
learning rate schedule & cosine decay  \\
warmup epochs & 40 \\
echo warmup epochs & 40 \\
training epochs & 1600 \\
gradient clip & 3.0 \\
label smoothing & 0.1 \\
dropout & 0.5 \\
masking ratio min & 0.5  \\
masking ratio max & 1.0  \\
masking ratio mode & 0.55 \\
masking ratio std & 0.25 \\
\hline
\end{tabular}
\caption{Pre-training Setting.}\label{tab:pretraining_settings}
\end{table}

\begin{table}[!htpb]
\centering
\begin{tabular}{|c|c|}
\hline
\textbf{config} & \textbf{value} \\
\hline
optimizer & LARS \\
base learning rate & 0.1 \\
weight decay & 0 \\
optimizer momentum & 0.9 \\
batch size & 4096 \\
learning rate schedule & cosine decay \\
warmup epochs & 0 \\
training epochs & 90 \\
augmentation & RandomResizedCrop \\
\hline
\end{tabular}
\caption{Linear Probing Setting.}\label{tab:linear_probing_settings}
\end{table}

\begin{table}[!htpb]
\centering
\begin{tabular}{|c|c|}
\hline
\textbf{config} & \textbf{value} \\
\hline
optimizer & LARS \\
base learning rate & 1.0 \\
weight decay & 0.0 \\
optimizer momentum & 0.9 \\
batch size & 16 \\
learning rate schedule & cosine decay \\
warmup epochs & 0 \\
training epochs & 10 \\
augmentation & RandomResizedCrop \\
\hline
\end{tabular}
\caption{Few-shot Setting.}\label{tab:few_shot_settings}
\end{table}

\section{MAGE Multi-setup Comparison}\label{multi_setup}
%\jesus{Improve name for god's sake}

Table \ref{tab:vs-mage} presents numerical results comparing MAGE and Sorcen, as visualized in Figure \ref{fig:magevssorcen}.  As discussed, Sorcen improves upon the state-of-the-art in unified learning, surpassed only by specific MAGE versions that lack balanced performance in both discriminative and generative tasks, thus losing their unified model advantage.

\begin{table}[!htbp]
\centering
\begin{tabular}{lrrr}
\toprule
Method & Epochs & FID & Linear \\
\midrule
MAGE & 800 & 11.60 & 73.30 \\
MAGE & 1600 & 11.10 & 74.70 \\
MAGE\textsubscript{wa} & 1600 & 8.67 & 70.50 \\
MAGE-C (1.0) & 800$^\dagger$ & 14.10 & 75.00 \\
MAGE-C (0.6) & 800$^\dagger$ & 27.00 & 77.10 \\
MAGE-C (0.6) & 1600$^\dagger$ & 31.77 & 78.20 \\ \midrule
Sorcen & 800 & 10.30 & 74.28 \\
\rowcolor[HTML]{9AFF99} Sorcen & 1600 & 9.61 & 75.10 \\
\bottomrule
\end{tabular}
\caption{Results on IN1k of different configurations of MAGE and Sorcen for 800 and 1600 pre-training epochs. $\dagger$ indicates the need for two passes of the VQGAN for each training step.}\label{tab:vs-mage}
\end{table}

\section{Additional Transfer Learning Results}\label{add_transfer}

In Tables \ref{tab:transfer_8shots} and \ref{tab:transfer_4shots} we further show the transfer learning performance of Sorcen in 8 and 4 shot setups respectively. While reducing the shots, the same behaviour maintains. Sorcen firmly beats MAGE on average for both setups.

\begin{table*}[!htpb]
    \centering
    \begin{tabular}{@{}lcccccccc@{}}
    \toprule
    Method        & Caltech & UCF101 & Flowers & Pets  & Sun   & EuroSAT & DTD   & Avg. \\ \midrule
    MAGE & 83.94   & 46.37  & \textbf{68.33}   & 48.79 & 43.38 & 40.01   & \textbf{40.13} & 52.99            \\
    Sorcen        & \textbf{85.31}   & \textbf{50.70}  & 67.37   & \textbf{50.89} & \textbf{44.65} & \textbf{46.70}   & 39.60 & \textbf{55.03}            \\ \bottomrule
    \end{tabular}
    \caption{\textbf{Transfer learning results (top-1 accuracy) for different datasets under 8-shot settings.} Last column contains the average across datasets.}\label{tab:transfer_8shots}
\end{table*}

\begin{table*}[!htpb]
    \centering
    \begin{tabular}{@{}lcccccccc@{}}
    \toprule
    Method        & Caltech & UCF101 & Flowers & Pets  & Sun   & EuroSAT & DTD   & Avg. \\ \midrule
    MAGE & \textbf{71.72}   & 29.71  & \textbf{46.00}   & 24.69 & 31.36 & 18.75   & 19.92 & 34.59            \\
    Sorcen        & 69.20   & \textbf{36.35}  & 43.89   & \textbf{24.91} & \textbf{33.48} & \textbf{25.01}   & \textbf{21.99} & \textbf{36.40}            \\ \bottomrule
    \end{tabular}
    \caption{\textbf{Transfer learning results (top-1 accuracy) for different datasets under 4-shot settings.} Last column contains the average across datasets.}\label{tab:transfer_4shots}
\end{table*}

\section{Precomputed MAGE}\label{precomputed}

\begin{table}[!htpb]
\centering
\begin{tabular}{lcc}
\hline
Method & IN1k val $^\dagger$ & IN200 val \\
\hline
MAGE$_{precomp}$ & 33.25 & 69.80 \\
\rowcolor[HTML]{9AFF99}Sorcen & \textbf{36.15} & \textbf{74.10} \\
\hline
\end{tabular}
\caption{\textbf{Results on precomputed token setups.} We highlight IN1k val with a $\dagger$ as it includes 800 new classes for the models, acting as new domain and showing the generalization capabilities of the models.}\label{tab:precomputed}
\end{table}

In Table \ref{tab:precomputed} we compare Sorcen with a modified version of MAGE that does not include the VQGAN inference during training. Therefore, MAGE is not able to apply any of the augmentations that it relies on. Both approaches are trained with the same precomputed IN200 dataset for 800 epochs and evaluated on IN200 validation and IN1k validation for 25-shot linear probing accuracy. As can be seen, MAGE falls further behind Sorcen as it was thought to be trained using, at least, Random Resized Crop augmentation which increases the diversity of the samples. This is also hinted in its Weak Augmentation setup (Table \ref{tab:vs-mage}), where it suffers a drastic drop on discriminative performance. Sorcen has been carefully designed to train on precomputed tokens and its Echo contrast provides required diversity to the framework. This diversity makes it surpass MAGE on IN1k validation set, which includes 800 new classes for the models.

\section{Disk Efficient Comparison}\label{disk}
During training, Sorcen does not require the VQGAN tokenizer. Every image is preprocessed beforehand and we work with the processed tokens. While our main goal consists in removing the overhead provided by the VQGAN \cite{esser_taming_2021}, we also reduce considerably the required disk space compared to the rest of the SoTA models. In Table \ref{tab:disk}, we compare Sorcen with disk-efficiency-focused models. To the best of our knowledge, Sorcen is the single disk efficient method with SoTA generation capacities. Using just the 0.29\% of the original dataset size, it outperforms previous disk efficient strategies, SeiT \cite{park_seit_2023} and competes with SeiT++ \cite{lee_seit_2025}, which uses dedicated token augmentations. Our approach, using less than a third of the space used by SeiT family \cite{park_seit_2023, lee_seit_2025}, manages to produce useful features that work for both discriminative and generative tasks, something impossible for SeiT family, which exclusively work on discrimination. Even if we do not focus on disk efficient learning, Sorcen opens an interesting line where 0.39GB of preprocessed data are enough to provide SoTA discriminative and generative results. This dataset will be released upon acceptance.

\addtolength{\tabcolsep}{-0.36em}
\begin{table}[t!]
\begin{tabular}{lllcc}
\toprule
Method                             & \# of images & Storage size & Top-1 & FID  \\ \midrule
MAGE  \cite{li_mage_2023}                             & 1,281k             & 140GB (100\%)        & 74.7       & 11.1 \\ \midrule
Uniform$^\dagger$            & 512k               & 54.6 GB (39\%)       & 74.0       & -    \\
C-score$^\dagger$ \cite{jiang_characterizing_2021} & 512k               & 53.3 GB (38\%)       & 73.3       & -    \\
JPEG 5$^\dagger$ & 1,281k             & 11.0 GB (8\%)        & 74.6       & -    \\ \midrule
SeiT    \cite{park_seit_2023}                           & 1,281k             & 1.4 GB (1\%)         & 74.0       & -    \\
SeiT++    \cite{lee_seit_2025}                         & 1,281k             & 1.4 GB (1\%)         & \textbf{77.8}       & -    \\ \midrule
\rowcolor[HTML]{9AFF99} Sorcen                                & 1,281k             & \textbf{0.39 GB (0.29\%)}     & \underline{75.1}       & \textbf{9.61} \\ \bottomrule
\end{tabular}
\caption{\textbf{Storage-efficient evaluation.} Linear accuracy and FID are provided. Sorcen is the single storage-efficient method with image generation capabilities. $\dagger$ results reported in \cite{park_seit_2023}.}\label{tab:disk}
\end{table}
\addtolength{\tabcolsep}{+0.36em}

\section{Hybrid Space Visualization}\label{sec:umaps}

Following other works in the literature \cite{estepa_all4one_2023, estepa2023good}, we use UMAP \cite{mcinnes2018umap} to visualize low-dimensional representations of the latent space of the model.
Thanks to this visualization, we can gain insight into how elements from different semantic classes are distributed in the representation space.

\begin{figure}[!htpb]
    \centering
    \begin{subfigure}[b]{0.495\linewidth}
        \centering
        \includegraphics[width=\linewidth]{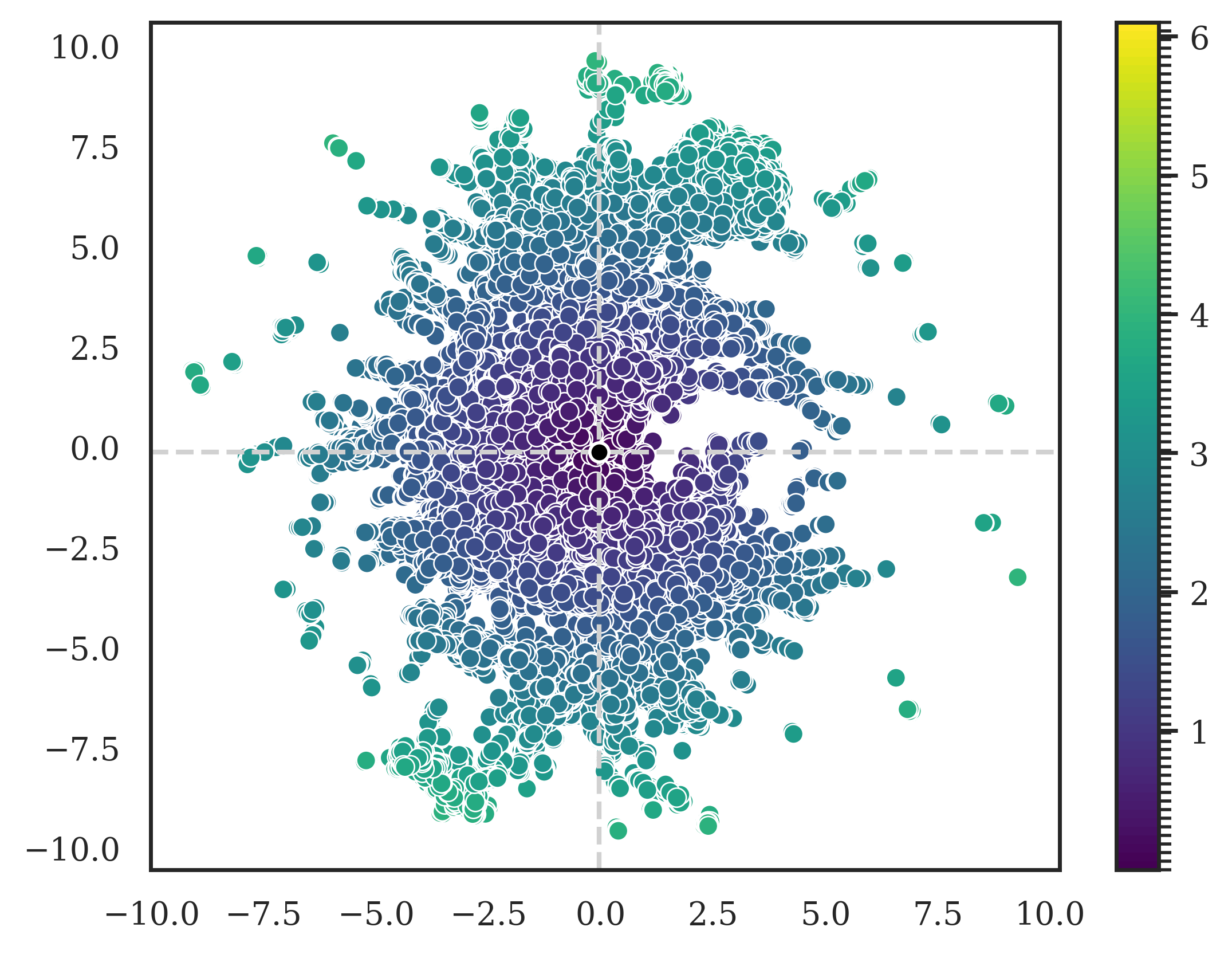}
        \caption{MAGE}
        \label{fig:sub1}
    \end{subfigure}%
    \begin{subfigure}[b]{0.495\linewidth}
        \centering
        \includegraphics[width=\linewidth]{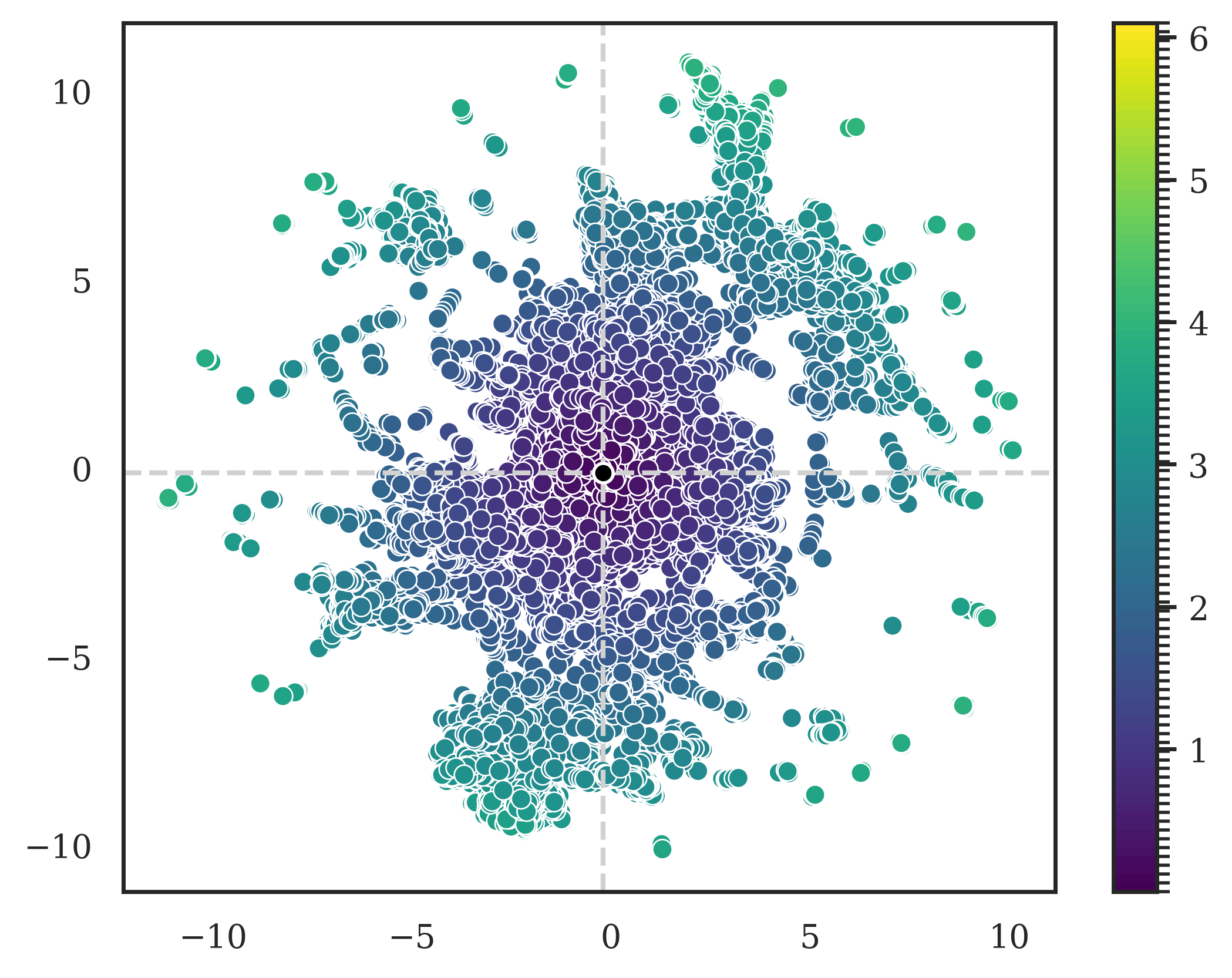}
        \caption{Sorcen}
        \label{fig:sub2}
    \end{subfigure}
    \caption{Latent space visualization (UMAP) of MAGE and Sorcen embeddings for the IN1k validation set. Each point represents an image, colored according to its distance (in standard deviations) from the mean embedding (black dot).}
    \label{fig:umap-general}
\end{figure}

\Cref{fig:umap-general} shows the UMAP representation of the whole IN1k validation set for both MAGE and Sorcen. According to this visualization, it is easy to see that, while the MAGE space resembles a \enquote{blob}, with almost no distinguishable clusters (apart from a few outliers), the space achieved by Sorcen presents easier-to-separate groups while preserving the dense nature of the MAGE space. This duality explains why Sorcen is able to excel in both discriminative and generative tasks by leveraging a common hybrid space.

\begin{figure*}[!htpb]
    \centering
    \includegraphics[width=\textwidth]{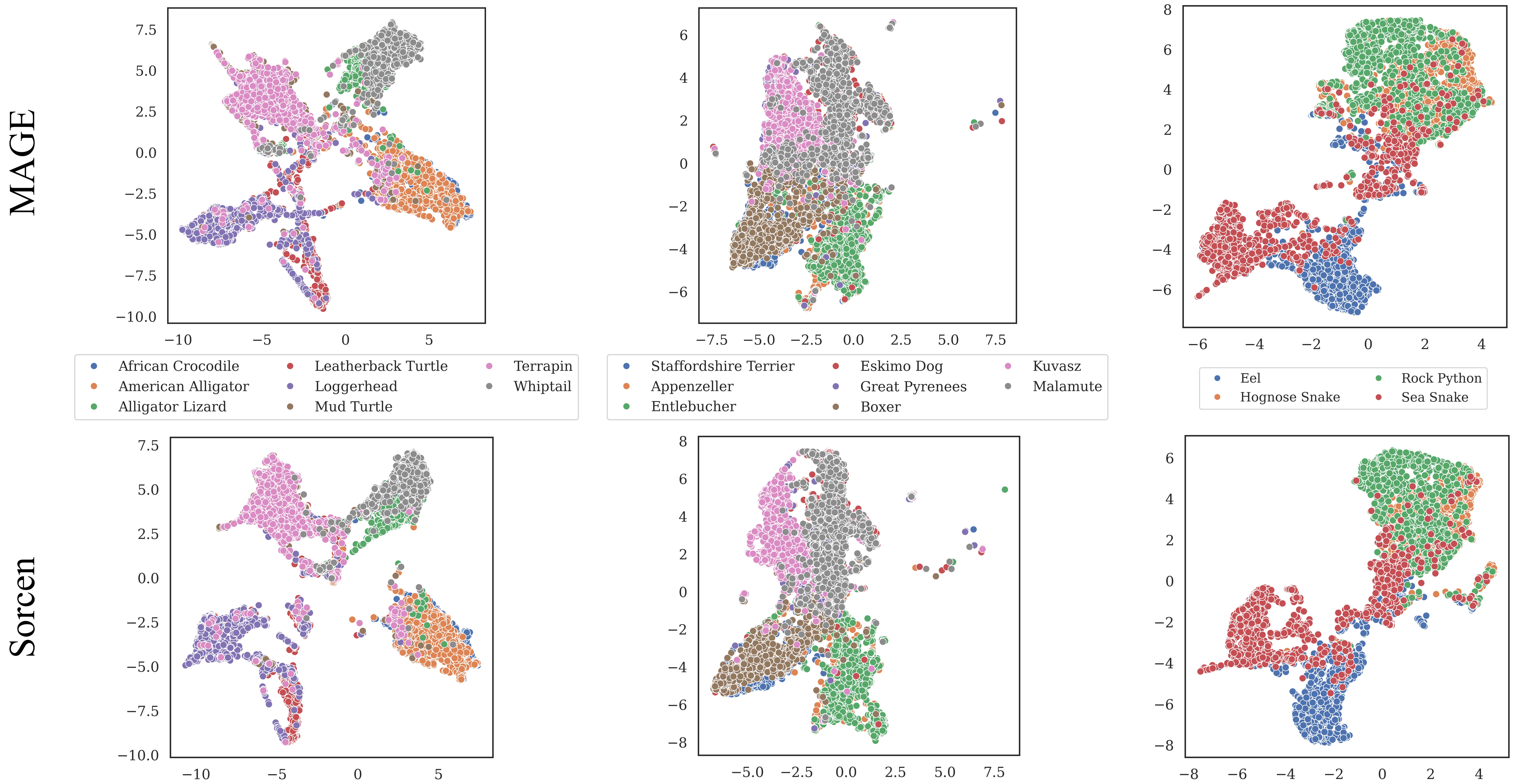}
    \caption{Class-specific UMAP visualizations of latent embeddings from MAGE (top row) and Sorcen (bottom row) for the IN1k training set. Points are colored by semantic class to illustrate class separation within each model's latent space.}\label{fig:umap-classes}
\end{figure*}

To further investigate the semantic organization within the latent space, we now visualize UMAP representations, color-coded by semantic class.  \Cref{fig:umap-classes} showcases these visualizations for a selection of representative classes from the IN1k training set, comparing both MAGE and Sorcen models. This class-based coloring allows us to examine the degree of separability between different semantic categories in each model's latent space, particularly for fine-grained categories.

Across the visualized classes, a consistent trend is observed: Sorcen's latent space exhibits markedly superior class separability compared to MAGE. While both models display some intra-class dispersion and inter-class mixing, this phenomenon is significantly more pronounced in MAGE. In Sorcen's representation, semantic classes are notably more distinct and less intermingled.

This enhanced class separability in Sorcen's latent space provides further insight into its hybrid capabilities.  The formation of more well-defined and separated class clusters likely underpins its stronger discriminative performance. Simultaneously, the preservation of an overall dense latent space, as demonstrated previously, still accommodates generative flexibility. Consequently, these class-based UMAP visualizations strongly suggest that Sorcen achieves a more semantically structured hybrid space. This structure effectively balances discriminative and generative representation learning, contrasting with MAGE's comparatively less organized latent space.

%\clearpage

\section{Impact of Hyperparameters}\label{sec:hyperparameters-values}

In this section, we delve into the impact of key hyperparameters on the performance of our Sorcen model. Through a series of ablation studies, we analyze the influence of the contrastive loss coefficient $\lambda$, the K parameter in top-k logit sampling, and the projector size. Due to computational constraints, we run these experiments on IN200 dataset for 200 epochs (1024 batch size) and evaluate them on few-shot regime with 25 shots per class. %The subsequent subsections, accompanied by Tables \ref{tab:coef}, \ref{tab:k}, and \ref{tab:proj}, detail our findings, demonstrating the importance of a well-configured contrastive loss coefficient and highlighting Sorcen's robust performance across a range of K and projector size settings.

\textbf{Contrastive loss coefficient. }We ablate the contrastive loss coefficient in Table \ref{tab:coef}. Consistent with prior work \cite{zhang_how_2022, huang_contrastive_2024, li_mage_2023}, reconstruction remains the dominant loss. However, reducing it below 0.1 inhibits the advantages provided by our echoed contrast, reducing the overall performance of the system. However, our ablation in Table \ref{tab:coef} reveals that the contrastive loss is essential.
Reducing the coefficient to 0.01 leads to a decrease in Top-1 accuracy to 66.77\% and a slight increase in FID to 22.60. In contrast, increasing the coefficient to 1.0 drastically degrades both metrics, with Top-1 accuracy falling to 47.77\% and FID increasing significantly to 46.82. This highlights that a balanced coefficient of 0.1 effectively balances discriminative and generative performance, achieving a good trade-off in Top-1 accuracy and a low FID score.

\begin{table}[t]
\centering
\begin{tabular}{@{}lccc@{}}
\toprule
     & 0.01  & 0.1   & 1.0   \\ \midrule
Top1 & 66.77 & 69.62 & 47.77 \\
FID  & 22.60     & 22.41     & 46.82     \\ \bottomrule
\end{tabular}
\caption{Study of the impact of the value of the contrastive loss coefficient $\lambda$ in accuracy (linear probing) and FID in IN200.}\label{tab:coef}
\end{table}

\textbf{K logit sampling. } 
Sorcen exhibits remarkable robustness to the K parameter in top-k logit sampling ask proved in Table \ref{tab:k}.
While larger K values introduce a potential risk of sampling less accurate tokens, the skewed nature of the logits produced by the decoder mitigates negative impacts. Conversely, smaller K reduces this risk but might also limit echo sample diversity.  Empirical results show negligible performance variation across K=5, 15, and 30, with Top-1 accuracy hovering around 69.6\% and FID consistently low (approximately 22.4-22.8).  We opted for K=15 as a balanced choice, mitigating inaccuracy risks while maintaining sufficient diversity in echo samples.
%Due to the skewed nature of the logits produced by the decoder, Sorcen proves to be robust regarding the K sampling parameter. While bigger K enables inaccurate tokens to be sampled by the distribution, the low probability of sampling them would reduce the negative impact of it. Lower K reduces this risk but it could also reduce the diversity of the echo samples. As the analysed results in Table \ref{tab:k} proved high robustness, we decided to keep the K value as 15 to balance inaccuracy risk and diversity.
%Across different K values, the model maintains a stable Top-1 accuracy around 69.6\%.  Similarly, the FID score remains consistently low, fluctuating slightly between 22.41 and 22.77.

\begin{table}[t]
\centering
\begin{tabular}{@{}lccc@{}}
\toprule
     & 5     & 15    & 30   \\ \midrule
Top1 & 69.66 & 69.62 & 69,60 \\
FID  & 22.69     & 22.41     & 22.77    \\ \bottomrule
\end{tabular}
\caption{Study of the impact of the number of tokens, K, considered to create the echos. Results correspond to IN200 pre-training and evaluation: linear probing and FID.}\label{tab:k}
\end{table}

\textbf{Projector size. } 
Sorcen achieves effective contrastive learning without requiring an excessively large projector.  Table \ref{tab:proj} shows that increasing projector size beyond 512 dimensions does not yield further performance gains.  In fact, expanding from 256 to 512 improves Top-1 accuracy (69.36\% to 69.62\%) and FID (22.74 to 22.41), indicating that 512 dimensions sufficiently capture relevant semantic information for contrast. However, further enlargement to 1024 and 2048 dimensions provides no significant advantage and even slightly degrades performance, suggesting diminishing returns and potential introduction of unnecessary model complexity beyond a projector size of 512.
%Sorcen does not require from a big projector output to succesfully contrast its pairs. In Table \ref{tab:proj} we show how increasing the projectors output size above 512 does not provide any performance boost. A projector size of 512 efficiently provides required semantics for Sorcens contrastive objective.
%Increasing the projector size from 256 to 512 improves the Top-1 accuracy from 69.36\% to 69.62\% and reduces the FID score from 22.74 to 22.41. However, further increasing the projector size beyond 512 does not yield substantial improvements and even slightly degrades the performance in terms of both Top-1 accuracy and FID score.

\begin{table}[t]
\centering
\begin{tabular}{@{}lcccc@{}}
\toprule
     & 256   & 512   & 1024  & 2048  \\ \midrule
Top1 & 69.36 & 69.62 & 69.55 & 69.45 \\
FID  & 22.74     & 22.41     & 22.52     & 22.19     \\ \bottomrule
\end{tabular}
\caption{Impact of the size of the projector in top-1 accuracy (linear probing) and FID in IN200. }\label{tab:proj}
\end{table}

\begin{comment}
\begin{table}[h]
\centering
\begin{tabular}{@{}lll@{}}
\toprule
             & Top-1 & Top-5 \\ \midrule
Sorcen$_{0.1}$  & 69.62 & 88.40 \\
Sorcen$_{0.01}$ & 66.77 & 87.30 \\
Sorcen$_{1.0}$  & 47.77 & 74.75 \\ \bottomrule
\end{tabular}
\caption{Loss coefficient ablation}
\end{table}
\end{comment}

\section{Qualitative Results}\label{qualitative}

Thanks to the SoTA image generation achieved by Sorcen, we are able to show in this section different generative applications of this model to modify and generate high-quality images. This can be done completely from scratch (unconditioned generation) or from a given part of an image. We also exhibit randomly selected Echos from Sorcens pretraining.
All the images shown in this section have been generated by Sorcen ViT-B trained for 1600 epochs on IN1k.

\begin{figure*}
    \centering
    \includegraphics[width=0.75\textwidth]{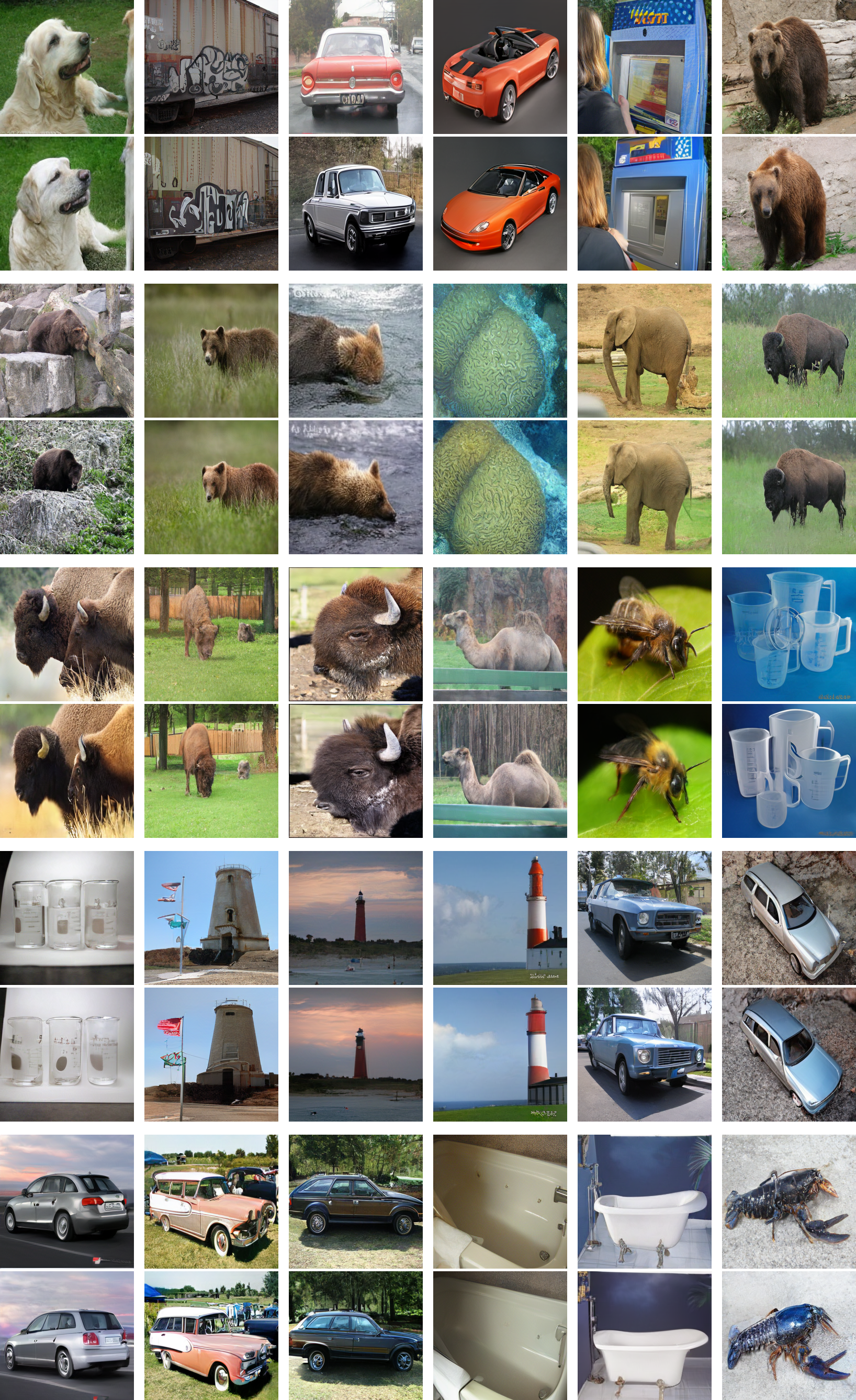}
    \caption{Uncurated collection of Echo samples generated during Sorcen training for contrasting. For each couple of rows, the top one contains the original image, and the bottom one the Echo.}\label{fig:more_echos}
\end{figure*}

\begin{figure*}
    \centering
    \includegraphics[width=0.75\textwidth]{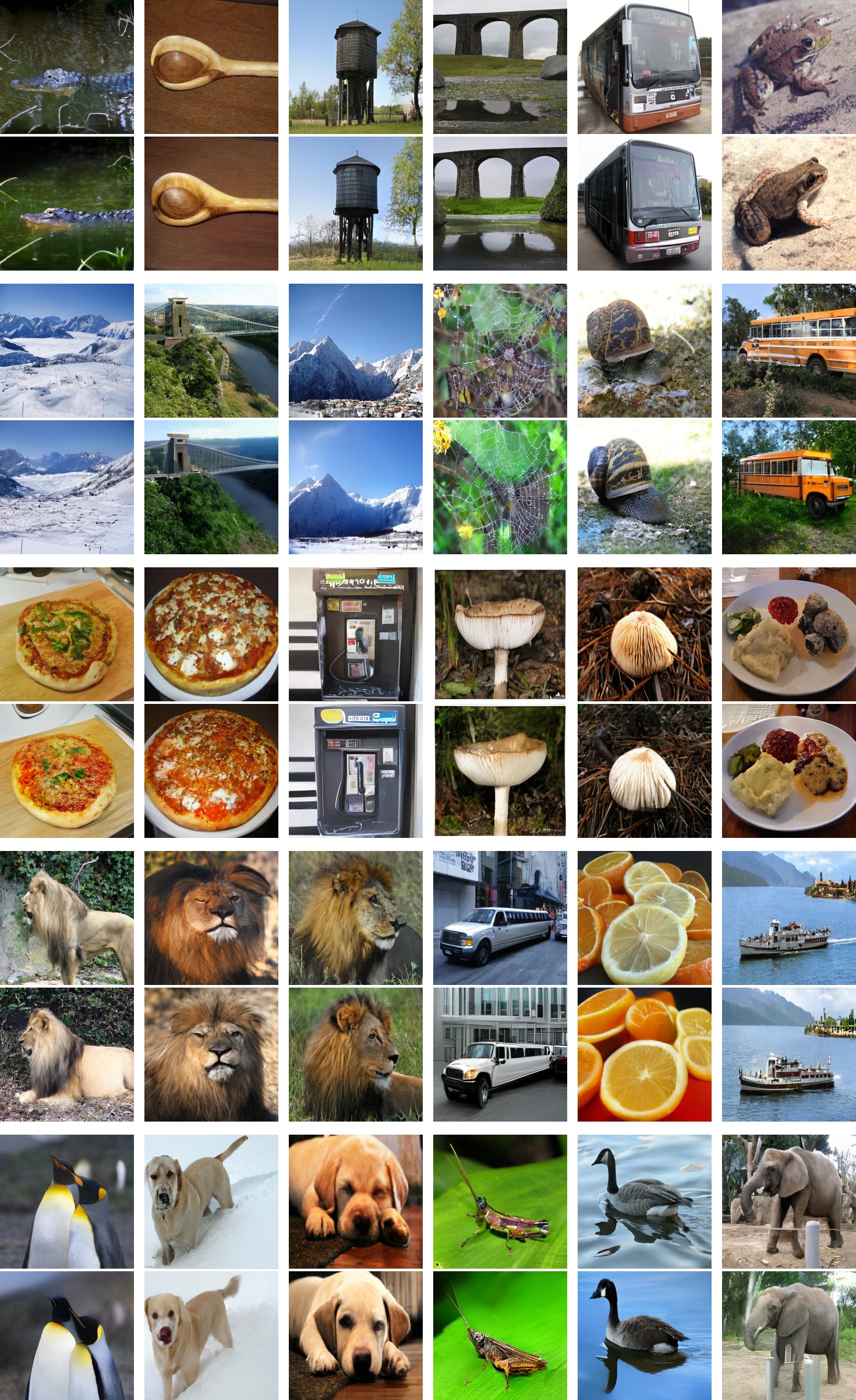}
    \caption{Uncurated collection of Echo samples generated during Sorcen training for contrasting. For each couple of rows, the top one contains the original image, and the bottom one the Echo.}\label{fig:more_echos_2}
\end{figure*}

\textbf{Echos extracted during training. }In Figures  \ref{fig:more_echos} and \ref{fig:more_echos_2} we display some of the Echos generated by Sorcen during training. 

\begin{figure*}
    \centering
    \includegraphics[width=0.9\textwidth]{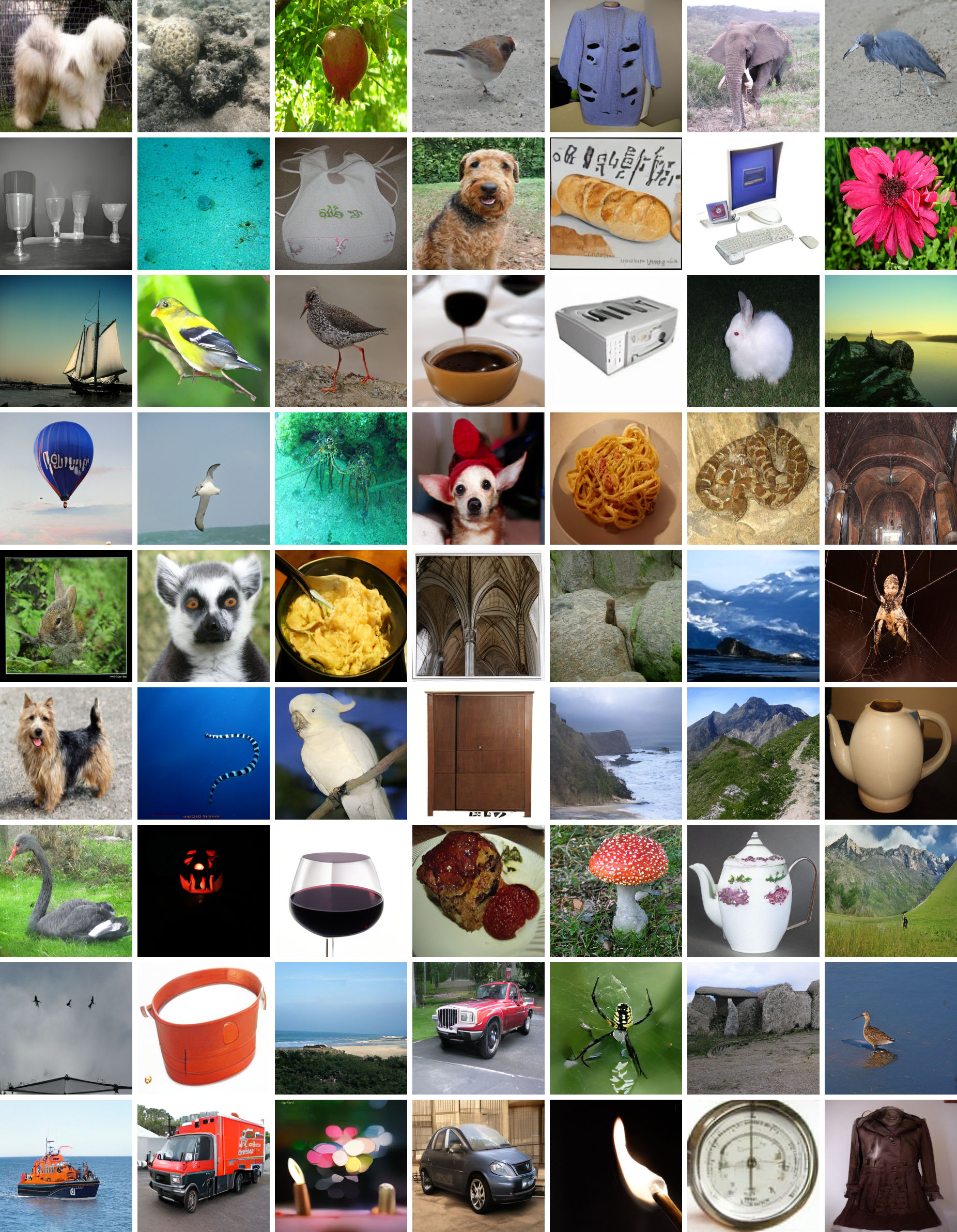}
    \caption{Images randomly generated by Sorcen in an unconditioned way using a ViT-B.}\label{fig:uncon}
\end{figure*}

\textbf{Unconditioned generation. } \Cref{fig:uncon} contains 63 examples of images randomly generated by Sorcen in an unconditioned way. 

\begin{figure*}
    \centering
    \includegraphics[width=0.9\textwidth]{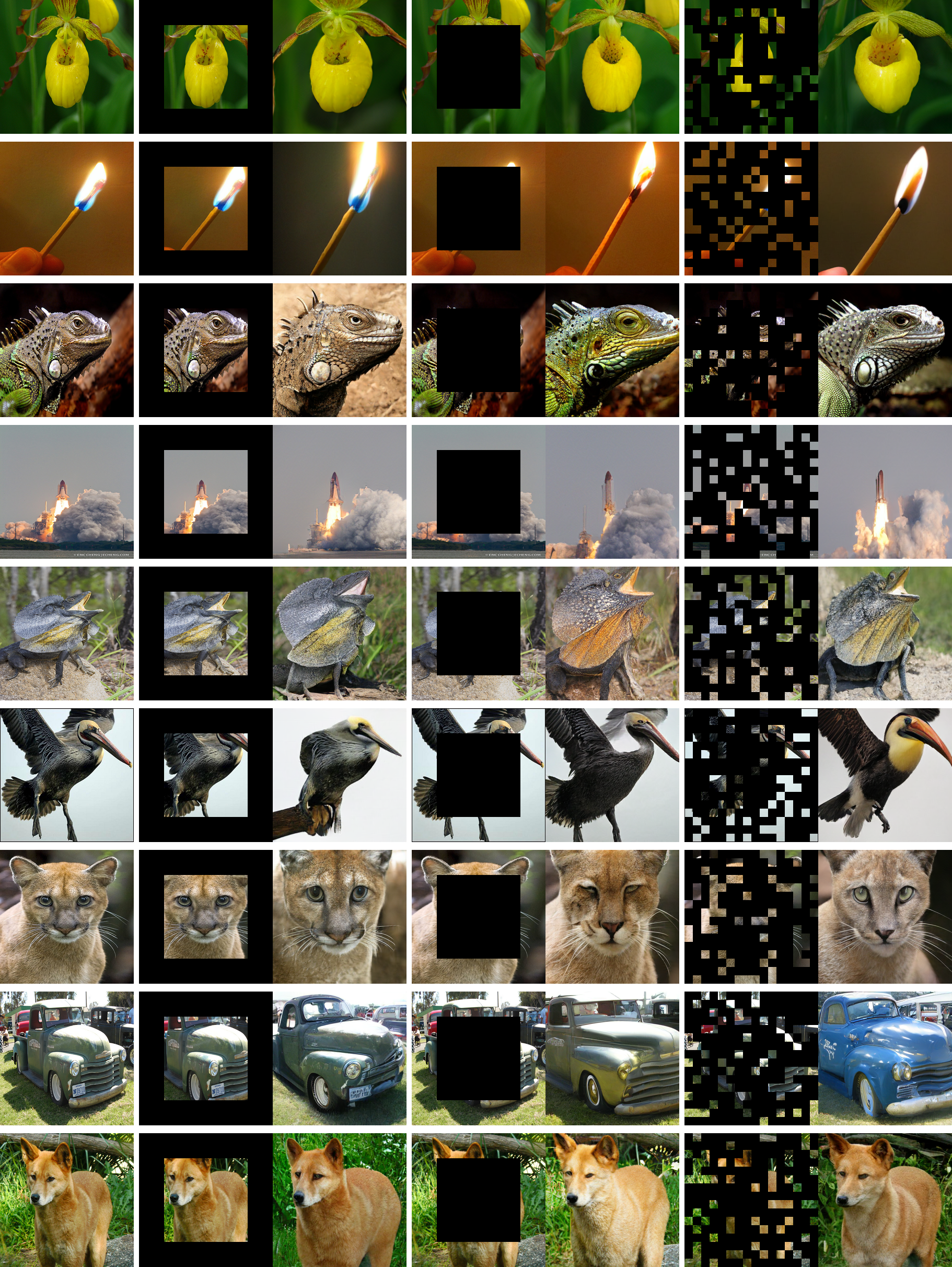}
    \caption{Examples of image inpainting, outpainting and reconstruction with Sorcen ViT-B. The first column is the original image.}\label{fig:inpainting1}
\end{figure*}

\begin{figure*}
    \centering
    \includegraphics[width=0.9\textwidth]{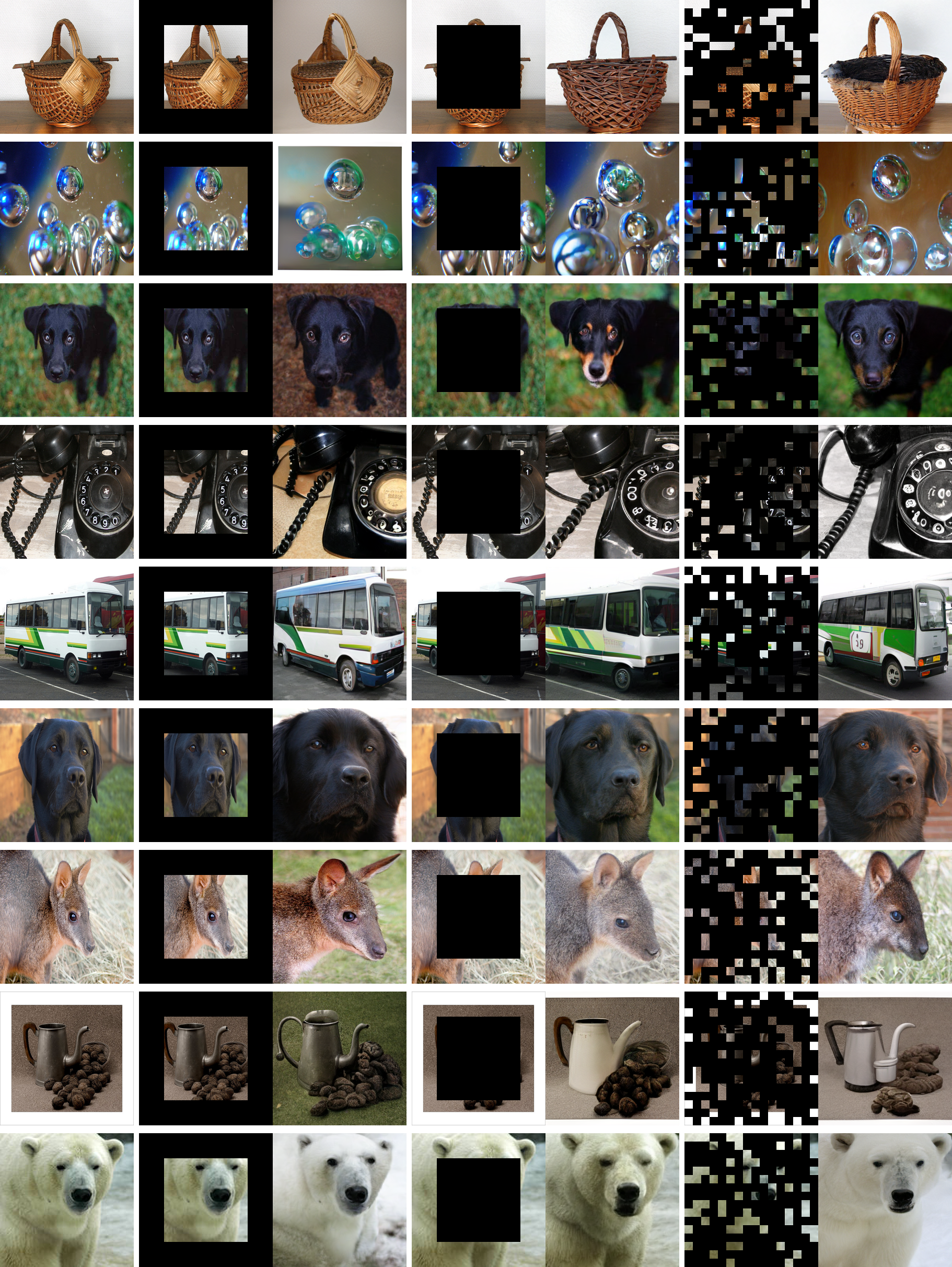}
    \caption{Examples of image inpainting, outpainting and reconstruction with Sorcen ViT-B. The first column is the original image.}\label{fig:inpainting2}
\end{figure*}

\begin{figure*}
    \centering
    \includegraphics[width=0.9\textwidth]{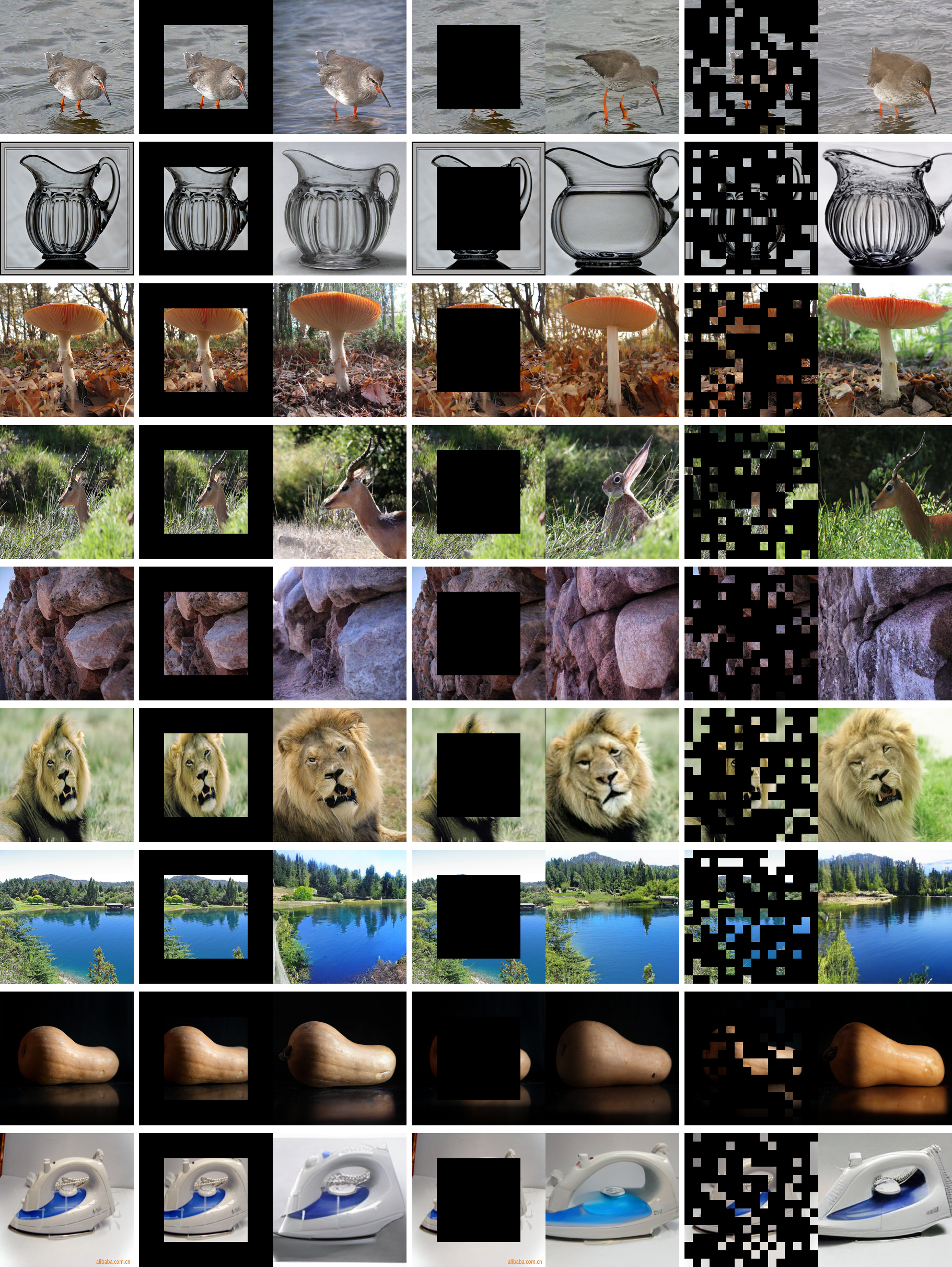}
    \caption{Examples of image inpainting, outpainting and reconstruction with Sorcen ViT-B. The first column is the original image.}\label{fig:inpainting3}
\end{figure*}

\textbf{Image inpainting/outpainting and reconstruction. } \Cref{fig:inpainting1,fig:inpainting2,fig:inpainting3} contain examples of inpainting, outpainting and reconstruction applied to images from IN1k dataset. Each rows correspond to an image. For reconstruction, we use a random mask of 75\% of the image. The first column is the original image, and the next columns contain pairs masked image-generation for three masks.

\clearpage

\begin{comment}
    \section{Rationale}
\label{sec:rationale}
% 
Having the supplementary compiled together with the main paper means that:
% 
\begin{itemize}
\item The supplementary can back-reference sections of the main paper, for example, we can refer to \cref{sec:intro};
\item The main paper can forward reference sub-sections within the supplementary explicitly (e.g. referring to a particular experiment); 
\item When submitted to arXiv, the supplementary will already included at the end of the paper.
\end{itemize}
% 
To split the supplementary pages from the main paper, you can use \href{https://support.apple.com/en-ca/guide/preview/prvw11793/mac#:~:text=Delete%20a%20page%20from%20a,or%20choose%20Edit%20%3E%20Delete).}{Preview (on macOS)}, \href{https://www.adobe.com/acrobat/how-to/delete-pages-from-pdf.html#:~:text=Choose%20%E2%80%9CTools%E2%80%9D%20%3E%20%E2%80%9COrganize,or%20pages%20from%20the%20file.}{Adobe Acrobat} (on all OSs), as well as \href{https://superuser.com/questions/517986/is-it-possible-to-delete-some-pages-of-a-pdf-document}{command line tools}.
\end{comment}

\end{document}